\documentclass[sigconf]{acmart}
\usepackage{subcaption}
\usepackage{multirow}
\usepackage{graphicx}
\usepackage{epstopdf}
\usepackage{fancyvrb}
\usepackage{tikz}
\usepackage{listings}
\usepackage{xcolor} 

\usetikzlibrary{arrows.meta, positioning, shapes.multipart}
\AtBeginDocument{%
  }

\acmConference[Agentic \& GenAI Evaluation at KDD 2025]{Agentic \& GenAI Evaluation Workshop at KDD 2025}{August 4, 2025}{Toronto, Canada}
\acmYear{2025}
\copyrightyear{2025}

\begin{document}

\title{Embeddings to Diagnosis: Latent Fragility under Agentic Perturbations in Clinical LLMs}

\author{Raj Krishnan Vijayaraj}
\email{raj.vijayaraj@alumni.utoronto.ca}
\affiliation{%
  \institution{Independent Researcher}
  \city{Toronto}
  \state{Ontario}
  \country{Canada}
}








\renewcommand{\shortauthors}{Vijayaraj}

\begin{abstract}

LLMs for clinical decision support often fail under small but clinically meaningful input shifts such as masking a symptom or negating a finding, despite high performance on static benchmarks. These reasoning failures frequently go undetected by standard NLP metrics, which are insensitive to latent representation shifts that drive diagnosis instability. We propose a geometry-aware evaluation framework, LAPD (Latent Agentic Perturbation Diagnostics), which systematically probes the latent robustness of clinical LLMs under structured adversarial edits. Within this framework, we introduce Latent Diagnosis Flip Rate (LDFR), a model-agnostic diagnostic signal that captures representational instability when embeddings cross decision boundaries in PCA-reduced latent space. Clinical notes are generated using a structured prompting pipeline grounded in diagnostic reasoning, then perturbed along four axes—masking, negation, synonym replacement, and numeric variation to simulate common ambiguities and omissions. We compute LDFR across both foundation and clinical LLMs, finding that latent fragility emerges even under minimal surface-level changes. Finally, we validate our findings on 90 real clinical notes from the DiReCT benchmark (MIMIC-IV), confirming the generalizability of LDFR beyond synthetic settings. Our results reveal a persistent gap between surface robustness and semantic stability, underscoring the importance of geometry-aware auditing in safety-critical clinical AI.

\end{abstract}





\maketitle

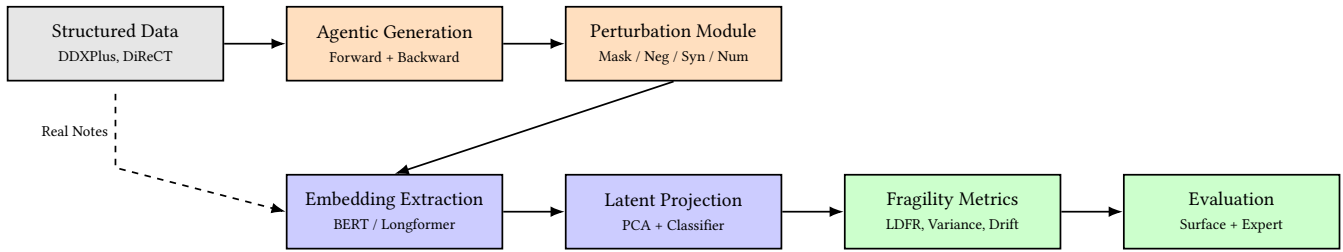
\begin{figure*}[t]
\centering
\resizebox{\textwidth}{!}{%
\begin{tikzpicture}[
  box/.style={rectangle, draw=black, thick, minimum width=3.5cm, minimum height=1.2cm, align=center},
  data/.style={box, fill=gray!20},
  gen/.style={box, fill=orange!25},
  embed/.style={box, fill=blue!20},
  eval/.style={box, fill=green!20},
  arrow/.style={-{Latex}, thick},
  node distance=1.5cm and 1.0cm
]

\node[data] (input) {Structured Data \\ \footnotesize DDXPlus, DiReCT};
\node[gen, right=of input] (gen) {Agentic Generation \\ \footnotesize Forward + Backward};
\node[gen, right=of gen] (perturb) {Perturbation Module \\ \footnotesize Mask / Neg / Syn / Num};

\node[embed, below=of gen] (embed) {Embedding Extraction \\ \footnotesize BERT / Longformer};
\node[embed, right=of embed] (pca) {Latent Projection \\ \footnotesize PCA + Classifier};
\node[eval, right=of pca] (metrics) {Fragility Metrics \\ \footnotesize LDFR, Variance, Drift};
\node[eval, right=of metrics] (eval) {Evaluation \\ \footnotesize Surface + Expert};

\draw[arrow] (input) -- (gen);
\draw[arrow] (gen) -- (perturb);

\draw[arrow] (perturb.south) -- (embed.north);

\draw[arrow] (embed) -- (pca);
\draw[arrow] (pca) -- (metrics);
\draw[arrow] (metrics) -- (eval);

\draw[arrow, dashed] (input.south) ++(0,-0.2) -- ++(0,-1.2) node[midway, left] {\footnotesize Real Notes} -- (embed.west);

\end{tikzpicture}
}
\caption{Overview of the LAPD evaluation pipeline. Synthetic or real clinical notes are processed through structured perturbation and latent embedding projection. The resulting representations are analyzed for fragility using geometry-aware metrics (e.g., LDFR) and surface-level clinical agreement.}
\label{fig:pipeline-tworow}
\end{figure*}

\section{Introduction}
LLMs like GPT-4\citep{gpt4} achieve near-human accuracy on medical benchmarks such as CMExam and MedQA\citep{medqa}, but these static metrics do not assess stability under clinically realistic variation. In high-stakes environments, robustness is very critical, like accuracy. While benchmarks provide a point estimate of correctness, they fail to measure whether models produce consistent diagnoses when inputs vary in natural but subtle ways.

Unlike output-level agreement metrics, LDFR reveals latent representation shifts that signal unstable diagnostic reasoning. Common metrics such as F1, EM, BERTScore\citep{bertscore}, BLEU\citep{bleu}, and ROUGE\citep{rouge} focus on surface similarity, not deeper clinical reasoning. They overlook shifts introduced by negation (e.g., “denies chest pain”), masking (e.g., privacy redactions), or synonym substitution (“heart attack” vs “myocardial infarction”). As a result, models may score highly yet remain brittle to changes that are common in clinical documentation.

We refer to this vulnerability as \textbf{diagnostic fragility}: when small, clinically grounded edits to input cause significant changes in the model’s output diagnosis. MedFuzz\citep{medfuzz} showed that even high-performing LLMs exhibit such volatility, but prior work has focused on whether the output label changes without examining how the internal representation shifts.

We argue that robustness also involves stability in the model’s internal reasoning. To capture this, we go beyond output agreement and introduce a metric that measures latent instability: the \textbf{Latent Diagnosis Flip Rate (LDFR)}. LDFR quantifies how often a structured perturbation causes the latent embedding to cross a diagnostic decision boundary in PCA-reduced space. These flips reveal underlying model sensitivity even when surface-level similarity remains high.

This motivates our framework: \textbf{LAPD (Latent Agentic Perturbation Diagnostics)}. LAPD uses synthetic clinical notes generated from an LLM-based pipeline and applies structured perturbations masking, negation, synonym replacement, and numerical edits at controlled thresholds. These edits are not semantically neutral, but deliberately designed as stressors to probe stability in diagnostic reasoning, as outlined in Figure~\ref{fig:pipeline-tworow}. To quantify this instability, we introduce the \textbf{Latent Diagnosis Flip Rate (LDFR)}, which measures how often perturbations cause embeddings to cross diagnostic decision boundaries in latent space.

In our analysis, we find that entity masking and negation consistently trigger large changes in LDFR, even when surface metrics like BERTScore remain above 0.9. Latent decision flips tend to occur along high-variance PCA axes, indicating that a small number of directions in the embedding space explain most of the instability. Importantly, these patterns generalize from synthetic notes to real clinical records from MIMIC-IV (DiReCT), confirming the broader applicability of our findings.

\noindent Our main contributions are:
\begin{itemize}
    \item We present \textbf{LAPD}, a geometry-aware evaluation framework that probes the latent robustness of clinical LLMs using structured, clinically grounded perturbations.
    \item We introduce the \textbf{Latent Diagnosis Flip Rate (LDFR)}, a model-agnostic metric that captures semantic instability through boundary crossings in PCA-reduced embedding space.
    \item We show that LDFR exposes diagnostic fragility overlooked by surface metrics, and generalizes to real clinical notes from DiReCT supporting its use in auditing clinical LLMs.
\end{itemize}

\begin{figure*}[t]
    \centering
    \includegraphics[width=0.9\linewidth]{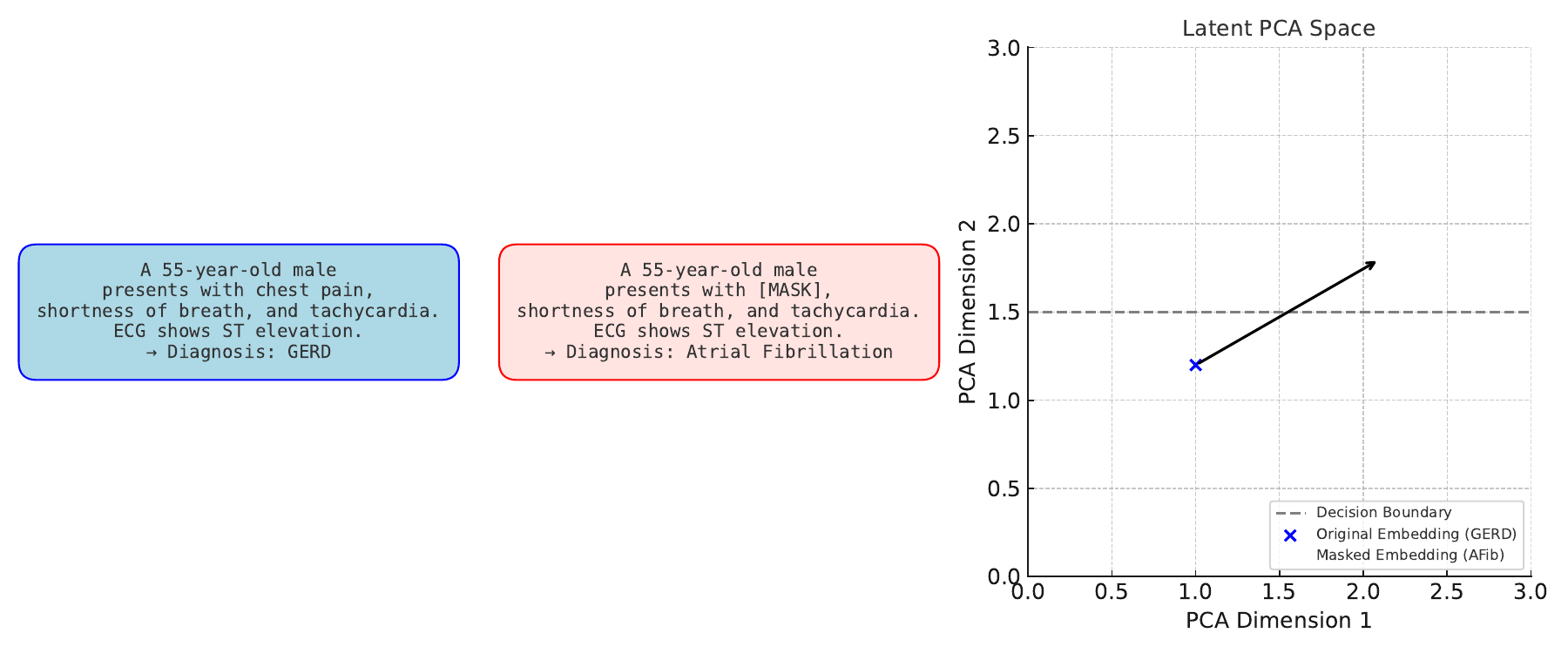}
\caption{
Overview of our framework (LAPD) through a constructed example illustrating latent diagnostic fragility. The left panel shows a synthetic clinical note generated by our prompting pipeline and diagnosed as GERD. The center panel applies masking to a key symptom (``chest pain''), resulting in a surface-similar variant that elicits a different diagnosis: Atrial Fibrillation. In the right panel, PCA-projected latent embeddings show that this perturbation crosses a diagnostic decision boundary. While this example is illustrative and not drawn from the evaluation set, it reflects a broader trend observed in our results: entity masking can cause latent representation shifts that lead to diagnosis flips, even under minor surface edits.
}
    \label{fig:initial_diagram}
\end{figure*}

\section{Related Work}

Robustness evaluation of clinical LLMs has primarily focused on surface-level stability under textual perturbations. Techniques such as MedFuzz \citep{medfuzz} and PerturbScore \citep{li-etal-2023-perturbscore} assess performance degradation under attribute-level fuzzing or paraphrase variations, but rely on output fidelity metrics like BERTScore or NER-F1. These approaches do not capture whether the underlying diagnostic reasoning process remains intact, nor whether latent representations remain within stable decision regions. Moreover, these evaluations are typically conducted on pre-curated datasets, rather than generated in an interactive, agent-driven pipeline.

Early representation methods such as Med2Vec \citep{Choi2016} and transformer-based EHR embeddings \citep{ehr_embedding} laid the groundwork for patient-level understanding, but they were not designed to capture robustness under naturalistic variation.

Recent work has explored geometry-aware metrics to analyze latent drift and embedding robustness. Studies have used PCA variance \citep{yin2018dimensionalitywordembedding}, Procrustes distance \citep{approx_lmi}, and low-rank subspace methods to characterize representation shifts. Manifold-based techniques \citep{ehr_manifold} and vocabulary unification strategies \citep{Johnson2024} have further aimed to align clinical semantics, though their robustness implications remain underexplored. The work by \citet{liu2025measuring} extends latent probing to adversarial reasoning stability by testing logical consistency properties such as transitivity and negation invariance across LLM outputs. However, their framework focuses on abstract reasoning tasks and does not ground latent shifts in specific clinical diagnostic decisions, as our LDFR metric does. Our proposed Latent Diagnosis Flip Rate (LDFR) fills this gap by detecting whether controlled, clinically plausible perturbations cause embeddings to cross decision boundaries tied to actual diagnostic outcomes. Complementary efforts in probing semantic generalization \citep{Pimentel2022OnTU} similarly explore latent-space reasoning, though not in diagnosis-specific settings.

In broader evaluation research, tools like ROSCOE \citep{roscoe}, ReCEval \citep{receval}, and CheckList \citep{checklist} test reasoning consistency and behavioral coverage, while factuality frameworks such as FactScore and FEVER assess correctness in generated content. TextFlint \citep{textflint} and TextAttack \citep{text_attack} provide adversarial input generation pipelines, but lack mechanisms to evaluate whether such perturbations lead to meaningful shifts in diagnostic decision-making. These methods are valuable for general-purpose LLM auditing, but are not designed to evaluate diagnostic integrity or label alignment in latent space. Unlike these, LDFR focuses on *reasoning stability* rather than hallucination detection or surface fluency.

A growing body of work now leverages agent-based frameworks not just for generation, but also for evaluation. \textbf{Agent-as-a-Judge framework} \citep{zhuge2024_agent_as_a_judge} proposes a process in which agents autonomously assess the reasoning accuracy of other agents' output, creating a scalable and automated audit loop. This paradigm inspires our own design, where the same LLM agents generate reasoning-grounded clinical notes and perform structured adversarial perturbations for robustness assessment. \citep{wang2025_latent_space_chain_of_embedding} introduces a latent chain probing strategy to assess LLM outputs using internal embedding dynamics rather than surface labels. Their findings support our core claim that latent shifts rather than just output deviations can serve as a diagnostic signal of reasoning instability. Our framework integrates these ideas by applying geometric probes to detect clinically-relevant diagnosis flips in a label-aware latent space. Related retrieval-augmented models like Med-PaLM \citep{medpalm} and clinical entity linking approaches \citep{clinical_entity} enhance factual grounding in LLMs, but remain largely unevaluated under perturbation.

Finally, the need for robust diagnostic evaluation arises across clinical NLP benchmarks. While tasks like MedQA \citep{medqa}, CMExam, and MedNLI focus on accuracy or entailment under static conditions, recent studies \citep{electronics13183781, li2025unravelinglocalizedlatentslearning} show that model representations in biomedical contexts are often unstable. Transformer-based clinical documentation systems \citep{chen-hirschberg-2024-exploring,Li2024ImprovingCN,wang2024direct} exhibit high performance on static tasks, but their internal consistency under adversarial conditions remains underexamined. LAPD unifies these strands by examining whether reasoning remains geometrically stable under adversarial perturbations.

\section{Agentic Generation and Perturbation Setup}

\subsection{Agentic Simulation of Diagnostic Notes}
The LAPD pipeline is shown in Figure~\ref{fig:pipeline-tworow}. Agentic note generation (left) begins with structured prompting grounded in DDXPlus, producing diagnosis-grounded narratives. Perturbation modules (middle) apply perturbation edits along four semantic axes. These perturbed notes are passed through frozen clinical encoders (right), and their latent representations are evaluated via PCA-based classification, variance analysis, and latent flip detection (LDFR).

We generate structured clinical notes using a reasoning-augmented agentic pipeline grounded in the DDXPlus dataset \citep{ddxplus}. Each note is produced through three chained prompts: (1) forward reasoning (symptoms to diagnosis), (2) backward justification (diagnosis to rationale), and (3) narrative construction. This agentic simulation creates interpretable, diagnosis-grounded synthetic notes while avoiding direct reuse of patient records. These controlled inputs enable systematic robustness evaluation. Full prompting templates are available in Appendix~\ref{appendix:prompts}.

\subsection{Clinically Grounded Perturbation Strategies}
We apply four types of clinically motivated perturbations to test robustness:
\begin{itemize}
    \item \textbf{Entity Masking (Omission):} Replaces medically salient entities with \texttt{[MASK]}. 
    \item \textbf{Negation (Omission):} Reverses polarity of symptoms, e.g., ``has chest pain'' $\rightarrow$ ``no chest pain.''
    \item \textbf{Synonym Replacement (Substitution):} Substitutes phrases with equivalent clinical terms.
    \item \textbf{Numerical Perturbation (Distortion):} Alters vitals/labs by \( \pm 5\%\text{--}15\% \).
\end{itemize}
Each method is applied at 0\%--100\% intensity by proportion of entities perturbed. For example, at 50\% threshold, half of all extractable clinical entities are modified. These edits are \textit{not} meant to preserve meaning, but to stress-test the LLM's internal diagnostic consistency.

\subsection{Evaluation on Real Clinical Notes}
To test robustness beyond synthetic data, we apply the same pipeline to 90 notes from the DiReCT benchmark based on MIMIC-IV. These notes average 870 tokens and include complex entity structures. We use Clinical-Longformer to encode full-length notes without truncation. This setting approximates real-world usage and complements our controlled synthetic evaluations.

\section{Latent Robustness Evaluation Framework}

\subsection{Surface-Level Evaluation Baselines}
We compute the following standard metrics to quantify textual fidelity:
\begin{itemize}
    \item \textbf{ROUGE-L, BERTScore (F1):} Lexical and semantic similarity.
    \item \textbf{NER F1, Jaccard Index:} Entity preservation.
    \item \textbf{Centroid Drift:} Euclidean distance between original and perturbed note embeddings.
\end{itemize}
These serve as baselines. However, they fail to capture semantic instability where diagnosis changes despite surface similarity.

\subsection{Latent Representation and PCA Probing}
We embed notes using frozen ClinicalBERT (synthetic) or Clinical-Longformer (real). PCA is applied to unperturbed embeddings to retain 90\% variance, yielding 30--40 principal components. A logistic regression classifier is trained on these projections to predict original diagnoses. This classifier models latent decision boundaries. It is used solely for probing, not as a substitute model.

To ensure classifier robustness, we validate performance via 10-fold cross-validation. Hyperparameters are reported in Appendix~\ref{appendix:hyperparams}.

\subsection{Latent Diagnosis Flip Rate (LDFR)}
We define LDFR as the proportion of perturbed samples whose latent representations cross the diagnostic boundary:
\[
\text{LDFR}(t) = \frac{1}{N} \sum_{i=1}^N \mathbb{I}[d_0^{(i)} \neq d_t^{(i)}]
\]
where \(d_0^{(i)}\) and \(d_t^{(i)}\) are classifier predictions for unperturbed and perturbed notes, respectively. We also compute \textbf{DFR}, the LLM's own diagnostic flip rate, to assess alignment with boundary-crossing behaviour.

\subsection{Complementary Latent Metrics}
We further interpret latent robustness through:
\begin{itemize}
    \item \textbf{Centroid Displacement:} Mean embedding shift across perturbations.
    \item \textbf{Per-Dimension Variance:} Encoder output dimensional variance reveals which directions drive instability.
\end{itemize}

\subsection{Clinician Evaluation of Diagnostic Realism}
Two clinicians evaluated five synthetic notes on structure, diagnostic clarity, and reasoning depth (0--3 scale). Scores ranged from 2.0--2.7. Qualitative feedback noted missing vitals, incomplete differentials, and unrealistic symptom presentation. This affirms the use of controlled perturbations to surface latent fragility. Reviewers were independent and unaffiliated with the project.

\begin{figure*}[t]
    \centering
    \begin{subfigure}[b]{0.45\textwidth}
        \includegraphics[width=\linewidth]{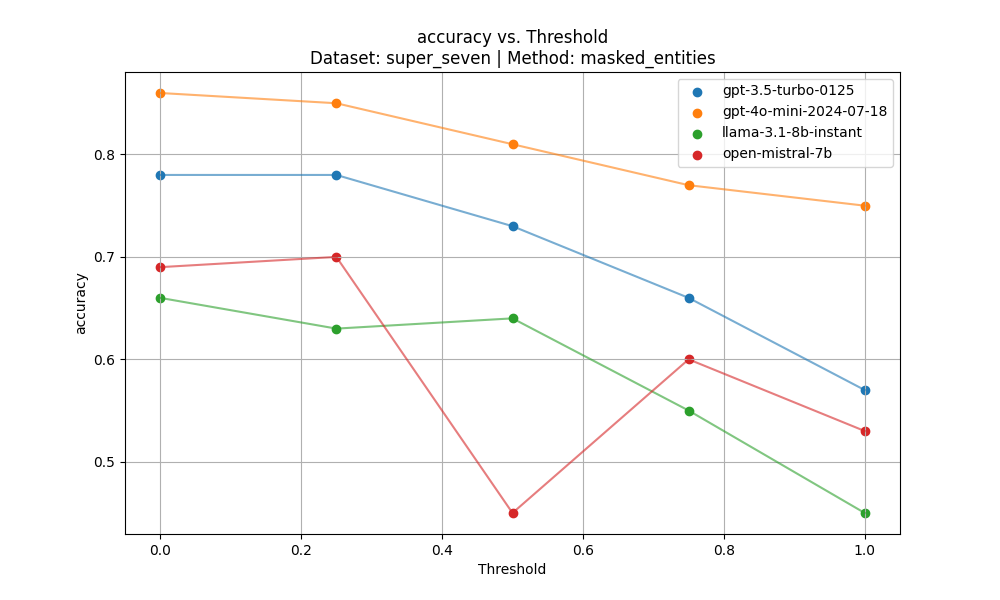}
        \Description{Line plot showing how diagnostic accuracy drops across increasing masking thresholds for each clinical LLM}
        \caption{Masked Entities: Most disruptive perturbation. Steep accuracy drop across all models suggests LLMs heavily rely on explicitly stated entities for diagnosis.}
        \label{fig:masked_accuracy}
    \end{subfigure}
    \hfill
    \begin{subfigure}[b]{0.45\textwidth}
        \includegraphics[width=\linewidth]{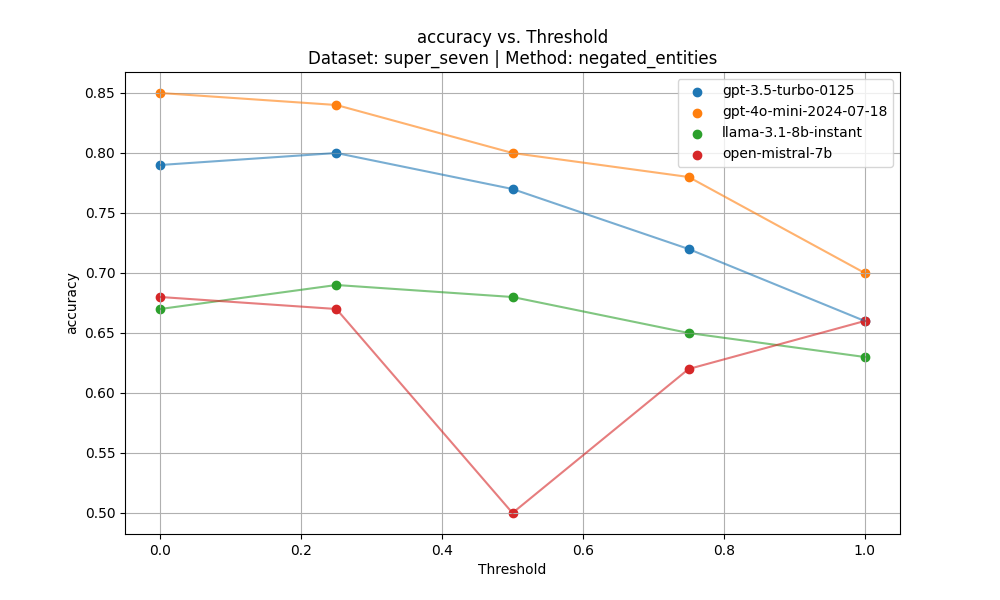}
        \Description{Line plot showing accuracy degradation as negation is applied to input notes for various LLMs.}
        \caption{Negated Entities: Moderate performance drop. Indicates partial sensitivity to polarity shifts like symptom presence vs. absence.}
        \label{fig:negated_accuracy}
    \end{subfigure}
    \vskip\baselineskip
    \begin{subfigure}[b]{0.45\textwidth}
        \includegraphics[width=\linewidth]{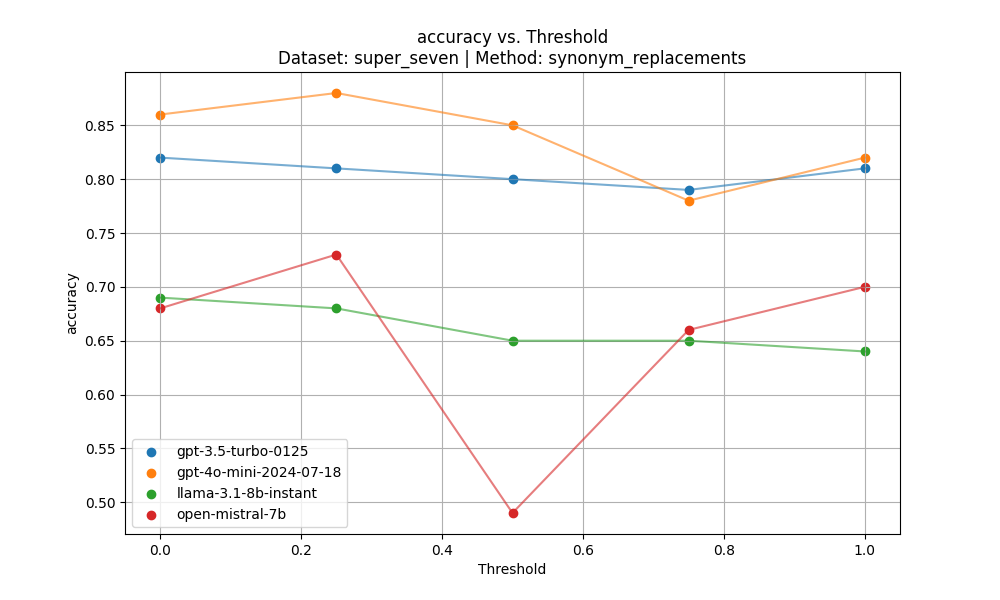}
        \Description{Line plot showing varied impact of synonym replacements on model accuracy across LLMs.}
        \caption{Synonym Replacements: Model-dependent effects. Some models show resilience while others misinterpret lexical variants.}
        \label{fig:synonym_accuracy}
    \end{subfigure}
    \hfill
    \begin{subfigure}[b]{0.45\textwidth}
        \includegraphics[width=\linewidth]{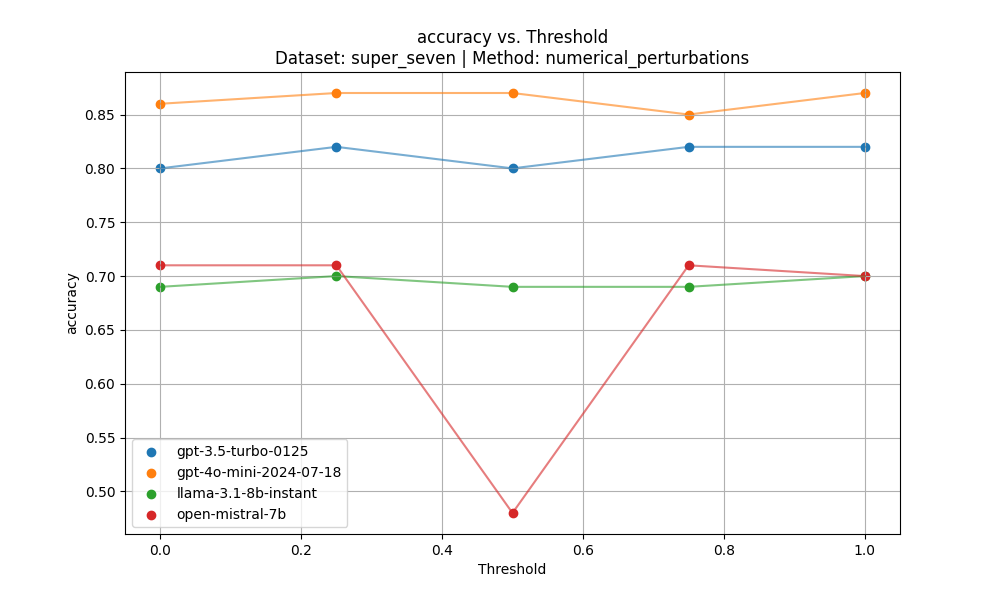}
        \Description{Line plot showing minimal accuracy drop with increasing numerical perturbations across LLMs.}
        \caption{Numerical Perturbations: Least disruptive. Accuracy remains mostly stable, suggesting LLMs underutilize quantitative signals in clinical text.}
        \label{fig:numeric_accuracy}
    \end{subfigure}
    \caption{LLMs show inconsistent resilience to clinically realistic perturbations. This figure illustrates how diagnostic accuracy degrades as we increase perturbation intensity (0--100\%) across four types. Masked entities cause the sharpest performance drop, while numerical edits have minimal effect, implying a greater LLM reliance on qualitative than quantitative reasoning.}
    \label{fig:accuracy_all}
\end{figure*}

\begin{figure*}[htbp]
    \centering
    \includegraphics[width=0.45\linewidth]{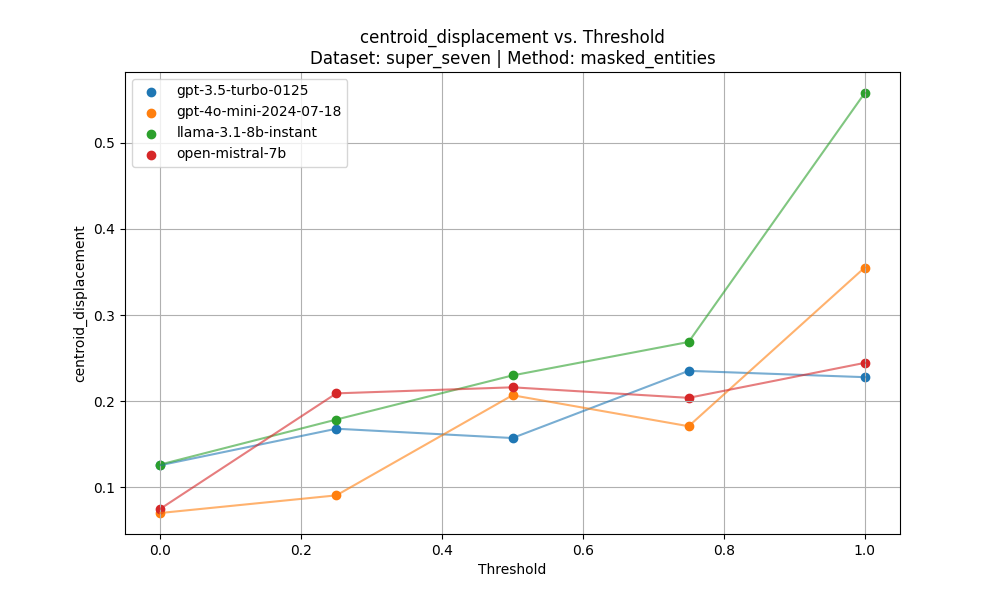}
    \includegraphics[width=0.45\linewidth]{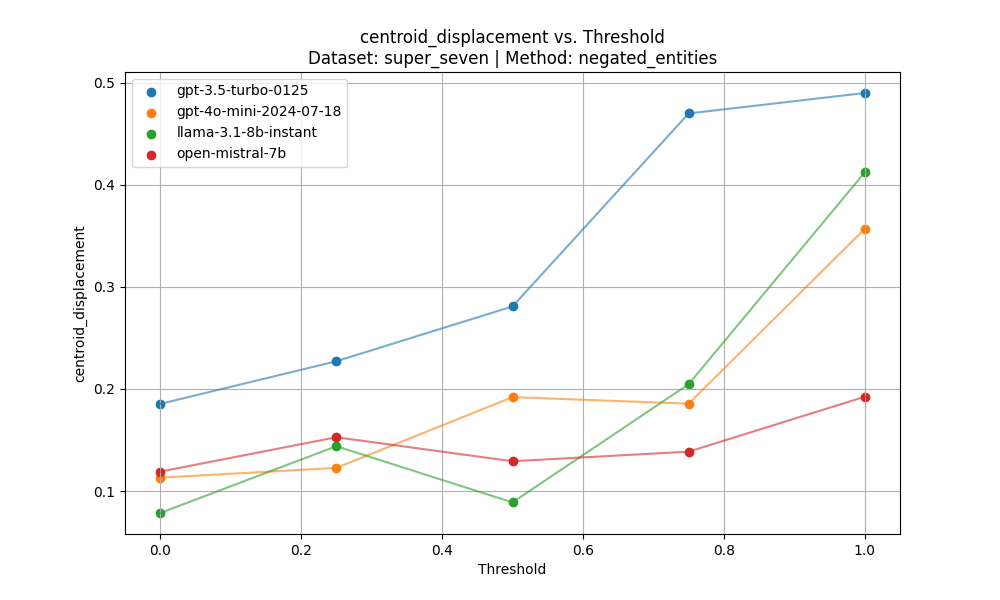}\\
    \includegraphics[width=0.45\linewidth]{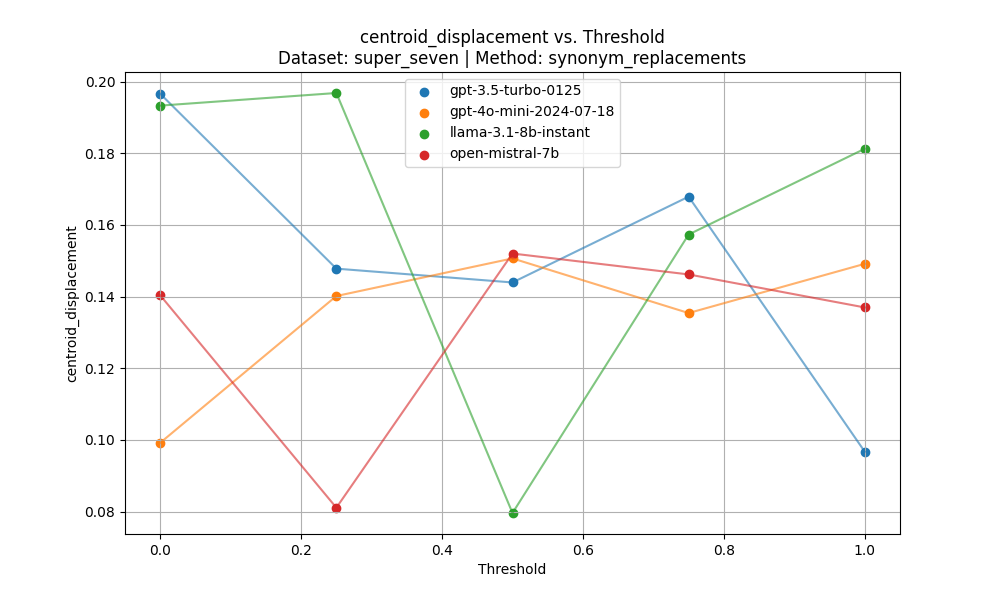}
    \includegraphics[width=0.45\linewidth]{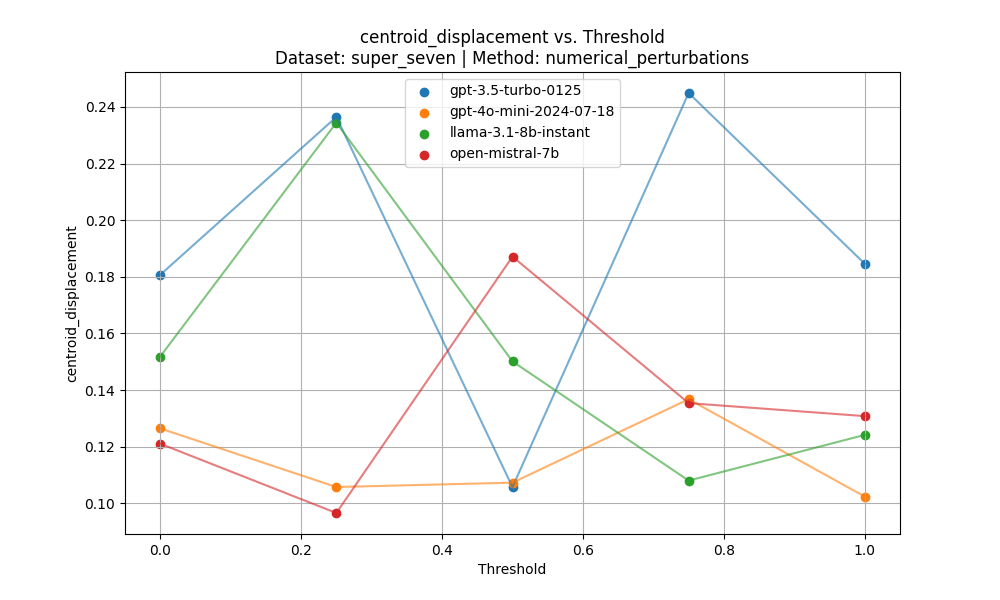}
    \Description{Four line plots showing centroid displacement across increasing perturbation thresholds (0 to 100 percent) for masked, negated, synonym, and numerical perturbations. Lines represent different clinical LLMs. Masking and negation cause large, monotonic shifts, while synonym and numeric perturbations are less consistent.}
    \caption{Global embedding shifts do not directly predict diagnostic failures. This figure shows how perturbations move embeddings away from their original centroid, measured by Euclidean distance. While masking and negation induce smooth, monotonic drift, these displacements are weakly correlated with diagnosis changes, suggesting that even small shifts can cross latent decision boundaries and lead to fragility.}
    \label{fig:centroid_displacement_grid}
\end{figure*}

\begin{figure*}[htbp]
    \centering
    \begin{subfigure}[b]{0.45\linewidth}
        \includegraphics[width=\linewidth]{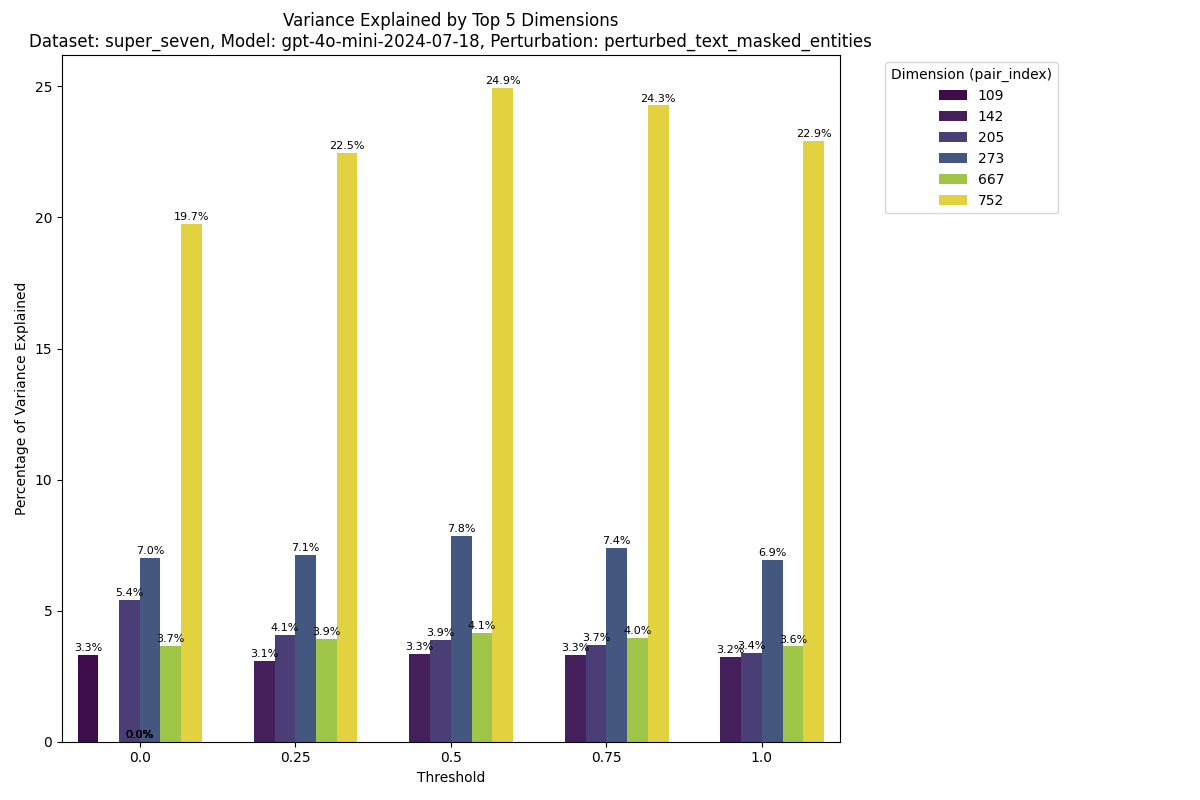}
        \caption{Masked Entities}
    \end{subfigure}
    \hfill
    \begin{subfigure}[b]{0.45\linewidth}
        \includegraphics[width=\linewidth]{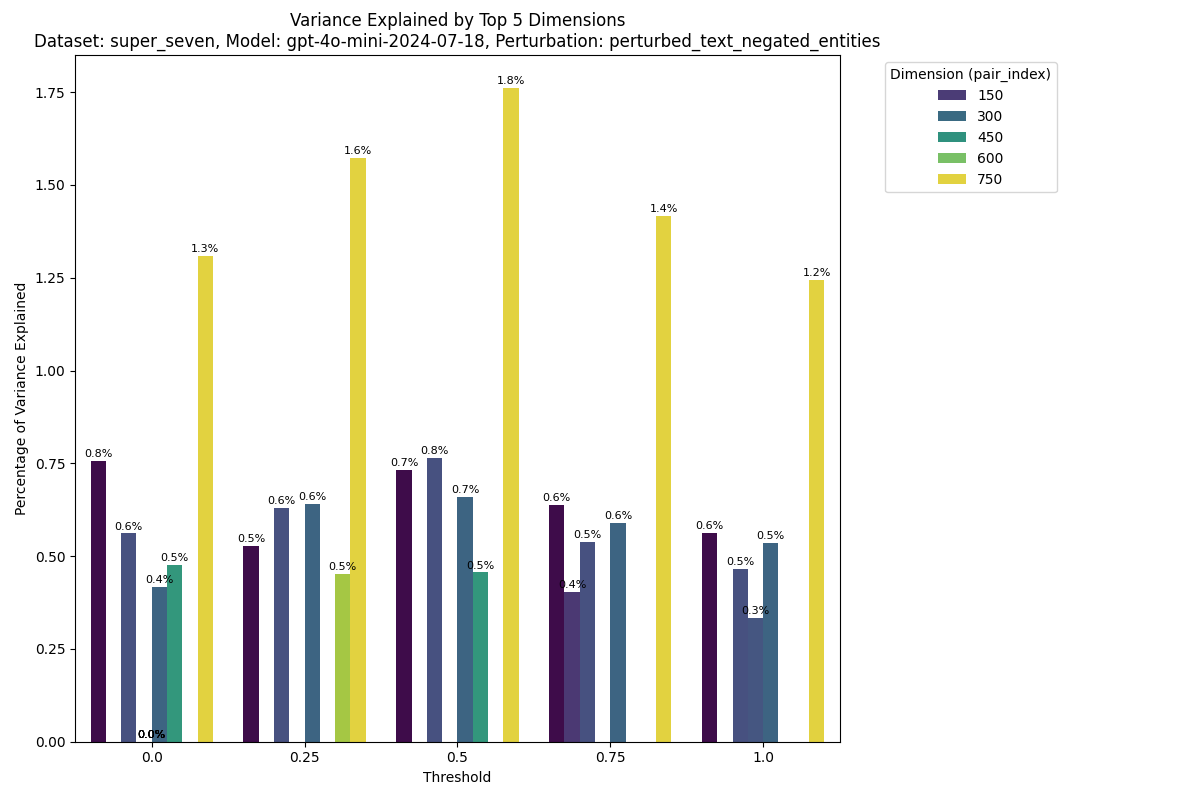}
        \caption{Negated Entities}
    \end{subfigure}
    
    \begin{subfigure}[b]{0.45\linewidth}
        \includegraphics[width=\linewidth]{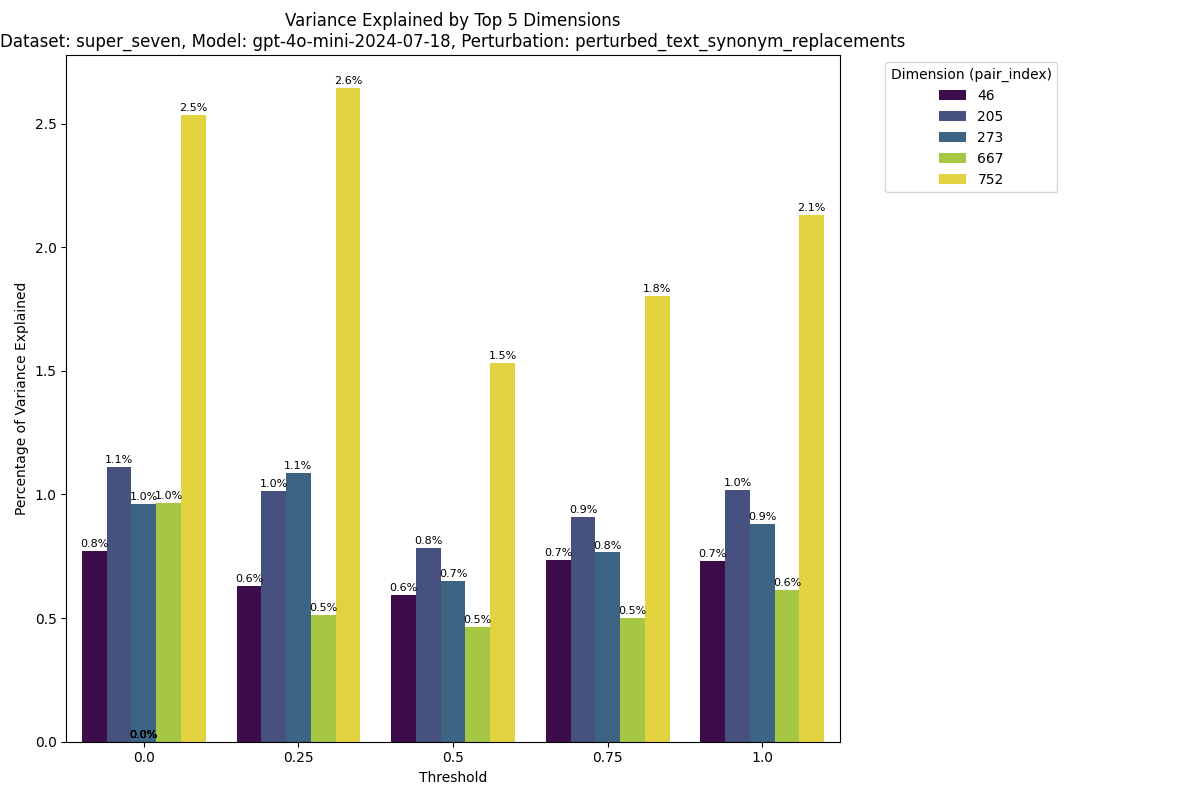}
        \caption{Synonym Replacements}
    \end{subfigure}
    \hfill
    \begin{subfigure}[b]{0.45\linewidth}
        \includegraphics[width=\linewidth]{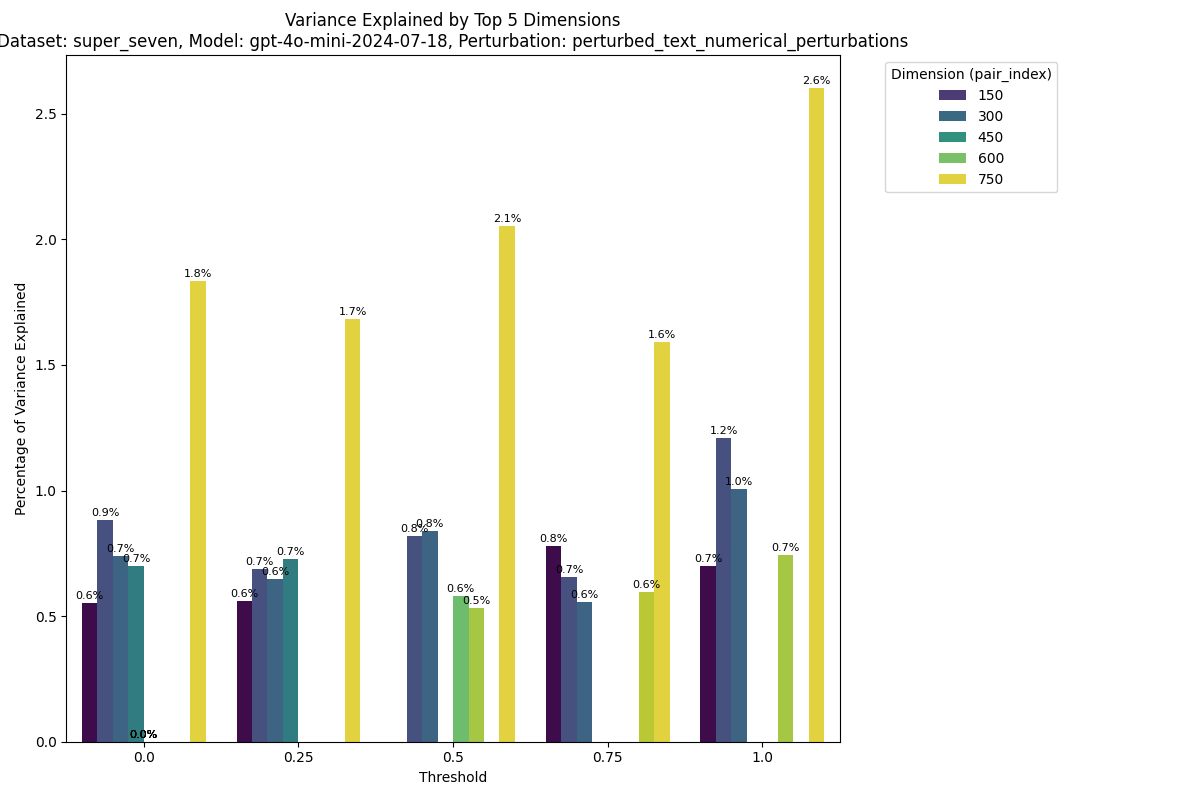}
        \caption{Numerical Perturbations}
    \end{subfigure}

    \Description{Four line plots showing the variance explained by the top 5 PCA dimensions of GPT-4O-mini embeddings under increasing perturbation thresholds, for masked, negated, synonym, and numerical changes. Masked entities concentrate variance in one or two dimensions, while numerical perturbations show flatter distributions.}
    \caption{Perturbations induce structured shifts in GPT-4O-mini’s latent space. This figure shows how variance is distributed across the top 5 PCA dimensions under each perturbation type. Masked entities cause variance to concentrate sharply along a few axes, indicating semantic bottlenecks, whereas numerical perturbations yield flatter, more distributed profiles. Such variance patterns reveal how fragility aligns with low-rank latent distortions.}
    \label{fig:dim_variance_gpt-4o-mini}
\end{figure*}

\begin{table*}[htbp]
\centering
\caption{Boundary crossings in latent space predict real diagnostic flips. This table reports Pearson and Spearman correlations between PCA-based classifier predictions and LLM outputs under increasing perturbation thresholds. Masked entities cause the sharpest drop in correlation, suggesting nonlinear semantic drift, while other perturbations (e.g., numerical) preserve alignment, highlighting varying degrees of latent instability.}
\Description{A table showing how well a PCA-based latent classifier aligns with LLM diagnosis outputs across four perturbation types—masked, negated, synonym, and numerical, measured by Pearson, Spearman, and average agreement with the LLM. Masked entities show a steep drop in correlation as perturbation intensity increases.}
\label{tab:ldfr_correlation}
\begin{tabular}{llccc}
\toprule
\textbf{Method} & \textbf{Threshold} & \textbf{Pearson} & \textbf{Spearman} & \textbf{LDFR} \\
\midrule
\multirow{5}{*}{Masked Entities}
  & 0.00 & 0.944 & 0.944 & 0.9125 \\
  & 0.25 & 0.493 & 0.493 & 0.5525 \\
  & 0.50 & 0.380 & 0.380 & 0.4900 \\
  & 0.75 & 0.322 & 0.322 & 0.4075 \\
  & 1.00 & 0.317 & 0.317 & 0.3525 \\
\midrule
\multirow{5}{*}{Negated Entities}
  & 0.00 & 0.944 & 0.944 & 0.9125 \\
  & 0.25 & 0.848 & 0.848 & 0.8525 \\
  & 0.50 & 0.744 & 0.744 & 0.8000 \\
  & 0.75 & 0.709 & 0.709 & 0.7375 \\
  & 1.00 & 0.702 & 0.702 & 0.7050 \\
\midrule
\multirow{5}{*}{Synonym Replacement}
  & 0.00 & 0.944 & 0.944 & 0.9125 \\
  & 0.25 & 0.845 & 0.845 & 0.8750 \\
  & 0.50 & 0.793 & 0.793 & 0.8400 \\
  & 0.75 & 0.728 & 0.728 & 0.7950 \\
  & 1.00 & 0.692 & 0.692 & 0.7325 \\
\midrule
\multirow{5}{*}{Numerical Perturbations}
  & 0.00 & 0.944 & 0.944 & 0.9125 \\
  & 0.25 & 0.894 & 0.894 & 0.8925 \\
  & 0.50 & 0.853 & 0.853 & 0.8750 \\
  & 0.75 & 0.900 & 0.900 & 0.8950 \\
  & 1.00 & 0.880 & 0.880 & 0.8925 \\
\bottomrule
\end{tabular}
\end{table*}

\begin{table*}[htbp]
\centering
\caption{LDFR alignment with LLM predictions on \textbf{real clinical notes} (DiReCT).}
\Description{Correlation between latent diagnostic flip rate and actual diagnostic flips across methods and thresholds, averaged over all models.}
\label{tab:ldfr_direct}
\begin{tabular}{llccc}
\toprule
\textbf{Method} & \textbf{Threshold} & \textbf{Pearson} & \textbf{Spearman} & \textbf{LDFR} \\
\midrule
\multirow{5}{*}{Masked Entities}
  & 0.00 & 0.5456 & 0.5456 & 0.6923 \\
  & 0.25 & 0.2640 & 0.2640 & 0.5187 \\
  & 0.50 & 0.3001 & 0.3001 & 0.5143 \\
  & 0.75 & 0.3377 & 0.3377 & 0.4681 \\
  & 1.00 & 0.2601 & 0.2601 & 0.4198 \\
\midrule
\multirow{5}{*}{Negated Entities}
  & 0.00 & 0.5456 & 0.5456 & 0.6945 \\
  & 0.25 & 0.1621 & 0.1621 & 0.4000 \\
  & 0.50 & 0.2991 & 0.2991 & 0.4110 \\
  & 0.75 & 0.2058 & 0.2058 & 0.3956 \\
  & 1.00 & 0.2130 & 0.2130 & 0.4198 \\
\midrule
\multirow{5}{*}{Synonym Replacement}
  & 0.00 & 0.5456 & 0.5456 & 0.6989 \\
  & 0.25 & 0.5270 & 0.5270 & 0.6879 \\
  & 0.50 & 0.4586 & 0.4586 & 0.6352 \\
  & 0.75 & 0.3406 & 0.3406 & 0.6022 \\
  & 1.00 & 0.4032 & 0.4032 & 0.6022 \\
\midrule
\multirow{5}{*}{Numerical Perturbations}
  & 0.00 & 0.5590 & 0.5590 & 0.6923 \\
  & 0.25 & 0.5270 & 0.5270 & 0.6857 \\
  & 0.50 & 0.5079 & 0.5079 & 0.6769 \\
  & 0.75 & 0.5079 & 0.5079 & 0.6989 \\
  & 1.00 & 0.4823 & 0.4823 & 0.6813 \\
\bottomrule
\end{tabular}
\end{table*}

\section{Experimental Setup}

We now describe the experimental setup used to instantiate the evaluation framework described in Section~4. This includes the choice of language models, dataset sources, perturbation statistics, and embedding details.

\subsection{Model Selection}
We evaluate two categories of models:

\textbf{General-Purpose Foundation Models:} \\ 
gpt-3.5-turbo-0125, gpt-4o-mini-2024-07-18, llama-3.1-8b-instant, and open-mistral-7b are included to assess fragility at the foundational level before domain-specific adaptation. This helps reveal vulnerabilities that may persist across deployment contexts.

\textbf{Clinical-Specific Models:} \\ 
\texttt{medgemma-27b} is tested to benchmark robustness among medical-domain fine-tuned models. These help evaluate if clinical training enhances perturbation resistance.

All models are accessed through inference APIs or local deployment, with temperature fixed at 0 or 0.01 and max token length of 512.

\subsection{Dataset Sources}
\textbf{Agentic Synthetic Notes:} \\ 
Derived from DDXPlus dialogues across 7 diagnoses: Pneumonia, Pulmonary Embolism, Atrial Fibrillation, Tuberculosis, GERD, Asthma, COPD. We generate 100 notes per diagnosis, each averaging 410 tokens (SD = 56) and containing ~8 extractable clinical entities.

\textbf{Real Clinical Notes:} \\ 
We evaluate 90 discharge summaries from the DiReCT benchmark (MIMIC-IV). These average 870 tokens (SD = 142) and contain ~13.2 named entities per note. This enables testing on longer, messier, and more realistic inputs.

\subsection{Perturbation Procedure and Matrix}
We apply four perturbation types (defined in Section~4.2) masking, negation, synonym replacement, and numerical edits, at 5 intensity thresholds (0\%, 25\%, 50\%, 75\%, 100\%).

\textbf{Perturbation Matrix Statistics:}
\begin{itemize}
  \item \textbf{Synthetic notes:} 100 unperturbed notes \( \times \) 4 perturbation types \( \times \) 5 thresholds \( = 2,000 \) perturbed samples.
  \item \textbf{Real notes:} 90 notes \( \times \) 4 types \( \times \) 5 thresholds \( = 1,800 \) additional samples.
\end{itemize}

We extract entities using a clinical NER pipeline, and perturbations are applied proportionally to threshold.

\subsection{Embedding and Classifier Configuration}
Synthetic notes are encoded using \texttt{ClinicalBERT}; real notes are encoded with \texttt{Clinical-Longformer} to avoid truncation. All embeddings are frozen. PCA is applied to unperturbed samples, retaining 90\% variance (typically 30–40 components). A logistic classifier is trained in this space to estimate latent boundaries.

Classifier performance is validated via 5-fold cross-validation, and hyperparameter details are in Appendix~\ref{appendix:hyperparams}.

\section{Results}
To assess the robustness of diagnostic reasoning in LLMs, we evaluate model behavior under structured perturbations across both surface-level and latent metrics. Our goal is to understand not only whether predictions change, but also how internal representations shift even when the text appears similar on the surface. We begin by comparing traditional NLP metrics with our latent diagnostic flip rate (LDFR) to uncover hidden fragility that surface scores may overlook.

\subsection{Surface vs Latent Metrics}

Appendix~\ref{appendix:nlp_metrics_figures} shows that surface metrics such as ROUGE-L and BERTScore remain high even when the model's diagnosis changes. For example, at 25\% masking, BERTScore stays above 0.89 while diagnosis accuracy drops significantly. At the same time, LDFR decreases, indicating that latent agreement breaks down even when surface similarity remains high. This highlights a key limitation of surface-level evaluation in clinical reasoning: it misses internal inconsistencies that lead to different predictions.

Figure~\ref{fig:accuracy_all} confirms this pattern across perturbation types. Masking causes the sharpest accuracy drop, while numerical changes have little effect, suggesting models may over-rely on surface cues and underuse numerical information. To better understand how this latent disagreement arises, we now examine the effect of specific perturbation types.

\subsection{Fragility by Perturbation Type}

Masked entities lead to the largest drop in both accuracy and LDFR as perturbation increases. This indicates strong dependence on explicitly stated symptoms. Negation causes a smaller but consistent drop in LDFR and accuracy, suggesting that polarity cues impact internal reasoning. Synonym and numeric changes maintain high ROUGE and BERTScore values, but still result in latent shifts, especially at higher perturbation levels. Table~\ref{tab:ldfr_correlation} summarizes these trends, showing how latent alignment degrades differently depending on the type of edit. Having established that perturbation types induce varied latent instability, we next ask: do different models handle this stress differently?

\subsection{Fragility by Model}

Model behavior under perturbation varies. GPT-4o and GPT-3.5-turbo show lower LDFR and smaller accuracy drops for synonym and numeric edits. In contrast, LLaMA-3.1 and Mistral are more fragile under masking and negation, with LDFR values exceeding 0.5 in some cases. These trends are visualized in Figure~\ref{fig:accuracy_all}, showing how different models react to increasing perturbation severity in accuracy. Notably, Mistral-7B, being a smaller and earlier-generation model, demonstrates poorer robustness, suggesting that certain diagnostic reasoning capabilities may emerge only at larger model scales.

\subsection{Latent Dimensional Collapse by Perturbation Type}
\label{sec:latent-collapse}

To complement global centroid drift, we analyze how variance redistributes across PCA dimensions under increasing perturbation (Figure~\ref{fig:dim_variance_gpt-4o-mini}). Under \textbf{masked entities}, up to 20–30\% of total variance is captured by a single dimension (e.g., dimension 752 in GPT-3.5 and LLaMA), indicating a bottleneck where embeddings compress into narrow subspaces. Negation and synonym replacements show model-specific trends: GPT-4o and Mistral maintain smoother variance distributions, while GPT-3.5 and LLaMA concentrate variance more tightly. Numerical perturbations produce flat variance curves across all models, reflecting stable and diffuse latent encodings.

These structured shifts align with our LDFR findings: as variance concentrates in fewer dimensions, LDFR increases, especially under masking and negation. Importantly, these effects occur within the 90\% variance PCA subspace (Appendix~\ref{appendix:pca_elbow}), preserving interpretability and classifier stability. In contrast, global centroid displacement (Figure~\ref{fig:centroid_displacement_grid}) increases smoothly but does not predict diagnostic changes. Together, these results suggest that perturbations expose low-dimensional instability patterns invisible to surface-level metrics.

\subsection{Real vs. Synthetic Notes}

\begin{table}[htbp]
\centering
\caption{Average tokens and entities per chunk for real and synthetic notes.}
\begin{tabular}{|l|c|c|}
\hline
\textbf{Type} & \textbf{Avg. Tokens} & \textbf{Avg. Entities} \\
\hline
DiReCT & 869.76 & 83.51 \\
synthetic\_open-mistral-7b & 369.97 & 30.16 \\
synthetic\_gpt-4o-mini-2024-07-18 & 533.28 & 27.49 \\
synthetic\_gpt-3.5-turbo-0125 & 328.36 & 31.98 \\
synthetic\_llama-3.1-8b-instant & 554.54 & 22.10 \\
\hline
\end{tabular}
\label{tab:avg_token_entities}
\end{table}

A key concern is whether our findings on synthetic notes hold for real clinical documentation. To test this, real clinical notes from the DiReCT dataset show similar patterns of latent fragility as synthetic notes. As shown in Table~\ref{tab:ldfr_direct}, correlations between LDFR and model predictions consistently drop as perturbations increase, confirming that our metric captures instability beyond synthetic settings.

Masked entity edits cause the largest decline in both cases, though the drop is more severe in synthetic notes. Notably, negation leads to sharper degradation in real notes, likely due to the complexity of natural language context. Synonym and numerical changes have smaller effects, with both note types showing similar trends. These results suggest that our framework generalizes across data sources and remains effective even on longer, noisier clinical text.

\subsection{Human Expert Validation}
\begin{table*}[ht]
\centering
\caption{Clinical expert evaluation of synthetic notes. Scores range from 0–3, where higher values indicate better note quality and diagnostic reasoning. Variability between reviewers highlights subjectivity in evaluating generated clinical text.}
\Description{Table showing expert evaluation scores for 5 synthetic clinical notes, rated on a scale of 0 to 3 by two reviewers for both note quality and diagnostic reasoning.}
\label{tab:clinician_eval}
\small
\begin{tabular}{ccccc}
\toprule
\textbf{Note ID} & \textbf{Note Quality (R1)} & \textbf{Note Quality (R2)} & \textbf{Reasoning (R1)} & \textbf{Reasoning (R2)} \\
\midrule
Note\_0 & 2 & 3 & 0 & 2 \\
Note\_1 & 2 & 3 & 2 & 3 \\
Note\_2 & 1 & 3 & 2 & 3 \\
Note\_3 & 3 & 3 & 3 & 3 \\
Note\_4 & 2 & 2 & 3 & 2 \\
\bottomrule
\end{tabular}
\end{table*}
Two clinicians scored five synthetic notes (0–3 scale). Mean scores were 2.0–2.7 across structure and diagnostic reasoning. Feedback highlighted missing vitals, incomplete differentials, and mismatches (e.g., hematemesis without GERD context). Table~\ref{tab:clinician_eval} summarizes scores. These qualitative insights affirm the need for latent audits even when surface fluency appears high.

\subsection{Diagnostic Auditing Potential}
LDFR identifies inputs where model outputs are unstable under light perturbations, flagging cases for downstream trust calibration. Unlike ROUGE or entity overlap, LDFR reveals semantic volatility invisible to surface metrics, offering a potential tool for auditing clinical assistant reliability. Taken together, our results show that LDFR captures subtle yet consequential breakdowns in diagnostic stability, providing a promising direction for future auditing tools in clinical AI systems.

\section{Conclusion}

We present a framework to uncover latent diagnostic instability in clinical language models by applying structured perturbations to synthetic clinical notes and analyzing semantic boundary shifts in latent space. Under entity masking, surface similarity remains high across thresholds, yet diagnosis flips become increasingly frequent highlighting fragility not captured by metrics like BERTScore. LDFR drops from 91.3\% to 55\% as perturbations increase, showing that the latent classifier no longer follows the LLM’s diagnosis reliably. This suggests that small changes in input can cause significant shifts in latent space, exposing both semantic drift and structural fragility. PCA-based boundary flips and per-axis variance shifts reveal latent vulnerabilities in diagnostic reasoning. Of all tested perturbations, entity masking proved most effective at exposing instability, inducing sharp transitions in low-dimensional latent space and misalignments that surface metrics fail to detect. Together, our findings underscore the need for geometry-aware evaluation to ensure safe, interpretable deployment of clinical LLMs in real-world settings. Our experiments with real notes from DiReCT and medical-specific models like MedGemma confirm that LDFR-based fragility patterns persist across synthetic and real clinical notes. Together, these findings support LDFR as a diagnostic tool for internal reasoning instability and a complement to traditional surface metrics in clinical LLM evaluation. Our central contribution, LDFR, offers a geometry-aware diagnostic signal that is not captured by surface-level metrics. Its consistent degradation under structured perturbations across both synthetic and real clinical notes highlights its utility for model auditing and evaluation beyond controlled benchmarks.

\section{Limitations and Future Directions}

Our framework offers a geometric perspective on diagnostic robustness but has limitations. Synthetic notes, while controllable, may not reflect real-world clinical variability, and using a fixed BERT embedding space may misalign with model-specific representations. While we include limited expert evaluation for unperturbed synthetic notes, we did not assess whether clinician reviewers agree with the diagnosis flips induced by perturbations. Future work should involve expert adjudication of perturbed samples to validate whether semantic edits, such as masking or negation, truly warrant diagnostic change. We selected a 90\% PCA variance threshold based on the elbow plot (Appendix~\ref{appendix:pca_elbow}), which showed that most diagnostic signal is captured within 30–45 components. This dimensionality preserves structure while ensuring that the LDFR classifier remains stable and interpretable across both synthetic and real note embeddings.

Future work will explore non-linear boundaries via manifold learning, expand perturbation analysis, and align latent shifts with expert judgments. Exploration of diagnostic manifolds as a foundation for clinical robustness is another direction illucidated.

\bibliographystyle{ACM-Reference-Format}
\bibliography{sample-base}

\appendix
\section{Appendix}
\subsection{Code and reproduction}
All prompting templates and code to reproduce our results are available at:  
\url{https://github.com/unni12345/geometric_diagnostics}

\subsection{The Prompts}
\label{appendix:prompts}
This prompt illustrated in Fig.\ref{fig:note_prompt} is designed to transform a structured patient–physician dialogue into a synthetic clinical note. The input includes socio-demographic details and question–answer pairs labelled by symptom or antecedent type. The prompt provides a well-defined clinical template with standard sections (e.g., Chief Complaint, History of Present Illness, Physical Exam), guiding the LLM to produce notes that are realistic, interpretable, and diagnostic-ready. This structure ensures that downstream perturbations can be applied in a controlled manner without compromising the core semantic structure.
\subsubsection{Clinical Note Prompt}
\begin{figure}[ht]
\centering
\begin{minipage}{0.9\linewidth}
\begin{lstlisting}
Patient Details:
Age: 44
Sex: F

Initial evidence provided: [{'question': 'Do you have pain somewhere, related to your reason for consulting?', 'is_antecedent': False, 'answer': True}]

Question & Answer Section:
For the question "Do you have pain somewhere, related to your reason for consulting?", 
the patient replied 'True', which is categorized as a symptom.
For the question "Characterize your pain:", the patient replied 'sickening', 
which is categorized as a symptom.
For the question "Do you feel pain somewhere?", the patient provided 'upper chest', 
indicating a symptom.

Clinical Report Template:
Based on the information provided, generate a comprehensive clinical report with the 
following sections:
- Patient Details
- Chief Complaint (extracted from the initial evidence)
- History of Present Illness: Describe onset, duration, severity, and relevant history.
- Past Medical History: Summarize any significant conditions.
- Medications and Allergies: List current medications and any known allergies.
- Physical Examination: Include vitals, general appearance, and pertinent exam findings.
\end{lstlisting}
\end{minipage}
\caption{Prompt for Clinical Note Generation.}
\label{fig:note_prompt}
\end{figure}

\subsubsection{Forward Reasoning Prompt}
This prompt illustrated in Fig.\ref{fig:forward_reasoning_prompt} initiates forward diagnostic reasoning: from clinical note to inferred diagnosis. The model is instructed to extract relevant observations and formulate logical deductions in a step-by-step manner. This chain-of-thought generation encourages explicit reasoning and provides transparency into the model's inference process. It is a crucial step for ensuring diagnostic traceability in the synthetic pipeline.

\begin{figure}[ht]
\centering
\begin{minipage}{0.9\linewidth}
\begin{lstlisting}
You are a diagnostic reasoning assistant. Read the following clinical note and generate 
a step-by-step reasoning process that extracts key observations and logical deductions 
leading to a diagnosis.

Clinical Note:
<-----Clinical Note----->

Provide the list of observations and deductions. Be concise.
\end{lstlisting}
\end{minipage}
\caption{Prompt for Forward Reasoning Generation.}
\label{fig:forward_reasoning_prompt}
\end{figure}

\subsection{Backward Reasoning Prompt}
This prompt illustrated in Fig.\ref{fig:backward_reasoning_prompt} facilitates backward reasoning: from an already inferred diagnosis to the supporting clinical evidence. The model must align symptoms, findings, and history in a way that supports the diagnosis logically. This step helps validate the consistency of the model's decision-making and serves as a diagnostic sanity check, reinforcing causal alignment between data and prediction.

\begin{figure}[ht]
\centering
\begin{minipage}{0.9\linewidth}
\begin{lstlisting}
You are a diagnostic reasoning assistant. Given the following clinical note and 
the final diagnosis '<diagnosis_placeholder>', generate a backward reasoning chain 
explaining how each observation supports the diagnosis.

Clinical Note:
<-----Clinical Note----->

Final Diagnosis: <diagnosis_placeholder>

Provide a concise list of observations and logical deductions.
\end{lstlisting}
\end{minipage}
\caption{Prompt for Backward Reasoning Generation.}
\label{fig:backward_reasoning_prompt}
\end{figure}

\subsubsection{Aggregator Prompt}
This prompt is illustrated in Fig.\ref{fig:aggregator_prompt} compares the forward and backward reasoning chains and compels the model to produce a final, consolidated rationale. The goal is to reconcile both perspectives into a coherent diagnostic explanation, reducing contradictions and highlighting mutually reinforcing observations. This step ensures logical integrity in the final output and simulates a clinician's process of reviewing both presentation and diagnostic hypothesis. 

\begin{figure}[ht]
\centering
\begin{minipage}{0.9\linewidth}
\begin{lstlisting}
You are a diagnostic reasoning aggregator. Compare the following two reasoning chains and 
generate the final, consolidated rationale behind the diagnosis:

Forward Reasoning:
<-----Forward Reasoning Chain----->

Backward Reasoning:
<-----Backward Reasoning Chain----->

Ensure that both chains are consistent in supporting the given diagnosis. Provide a concise 
list of key observations and deductions.
\end{lstlisting}
\end{minipage}
\caption{Prompt for Aggregation and Verification of Reasoning Chains.}
\label{fig:aggregator_prompt}
\end{figure}


\subsection{Note-to-Dialogue Comparison via Surface-Level Metrics}
\label{appendix:semantic_metrics}

For completeness, we report automatic evaluation scores comparing the generated clinical notes to their source DDXPlus dialogues using standard surface-level metrics. These include BERTScore, ROUGE-L, and biomedical NER-based Jaccard and F1 scores, summarized in Table~\ref{tab:super_seven_metrics}.

The absolute values of these metrics are relatively low (e.g., BERTScore ranges between 0.54–0.58), which is expected given the nature of the comparison: the source DDXPlus inputs are unstructured, sparse, and conversational, while the generated notes are formal and structured. As a result, high token or span level overlap is neither expected nor necessarily desirable.

We used bert-large-uncased for BERTScore and the 
d4data/biomedical-ner-all model for entity-level scoring. Both are pre-trained general-purpose tools and not specifically optimized for dialogue-to-note mapping. Furthermore, BERTScore captures token-level similarity and does not reflect higher-level clinical discourse structure or reasoning coherence.

These scores are therefore included for completeness and relative comparison across models, but they do not directly evaluate the quality of diagnostic reasoning or factual consistency. In fact, high surface-level similarity could indicate shallow copying, whereas meaningful abstraction or reasoning may naturally result in lower overlap—motivating the need for robustness evaluation beyond traditional text metrics.


\begin{table}[htbp]
\centering
\caption{Surface-level comparison between generated clinical notes and their source DDXPlus dialogues. Metrics include BERTScore, ROUGE-L, and biomedical NER-based Jaccard and F1 scores. The best-performing model per metric is bolded. While Open-Mistral-7B ranks highest on BERTScore and ROUGE-L, GPT-3.5-turbo yields the best entity-level scores—indicating that surface similarity and clinical entity preservation do not always align.}
\Description{Table comparing four language models—GPT-3.5-turbo-0125, GPT-4o-mini, LLaMA-3.1-8b, and Open-Mistral-7B—on BERTScore, ROUGE-L, NER Jaccard, and NER F1. Open-Mistral-7B scores highest on BERTScore and ROUGE-L, while GPT-3.5-turbo achieves the best NER metrics.}
\label{tab:super_seven_metrics}
\resizebox{\linewidth}{!}{%
\begin{tabular}{lcccc}
\toprule
\textbf{Model} & \textbf{BERTScore} & \textbf{ROUGE-L} & \textbf{NER Jaccard} & \textbf{NER F1} \\
\midrule
GPT-3.5-turbo-0125     & 0.546 & 0.221 & \textbf{0.171} & \textbf{0.283} \\
GPT-4o-mini-2024-07-18 & 0.559 & 0.219 & 0.141 & 0.243 \\
LLaMA-3.1-8b-instant   & 0.575 & 0.240 & 0.161 & 0.270 \\
Open-Mistral-7B        & \textbf{0.580} & \textbf{0.244} & 0.164 & 0.276 \\
\bottomrule
\end{tabular}%
}
\end{table}

\subsection{Supplementary Metric Plots}
\label{appendix:nlp_metrics_figures}

We have included full degradation plots for all NLP evaluation metrics—NER F1, ROUGE-L, and NER Jaccard—under each perturbation type. These figures provide a comprehensive view of how different perturbations affect semantic fidelity and entity preservation across clinical LLMs.

\subsubsection{NER F1 vs. Threshold}

\begin{figure*}[htbp]
    \centering
    \begin{subfigure}[b]{0.45\textwidth}
        \includegraphics[width=\linewidth]{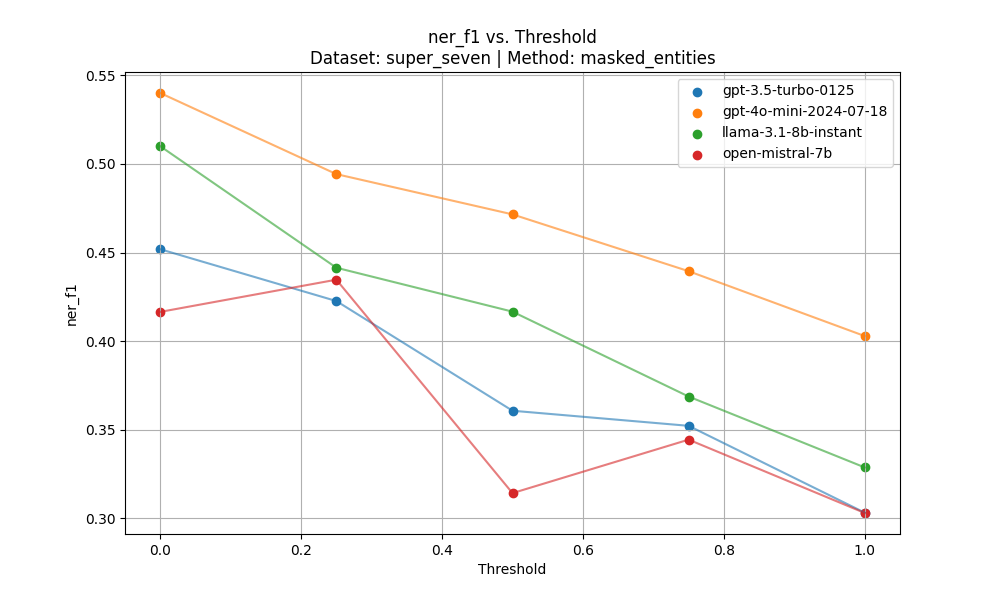}
        \caption{Masked Entities}
    \end{subfigure}
    \hfill
    \begin{subfigure}[b]{0.45\textwidth}
        \includegraphics[width=\linewidth]{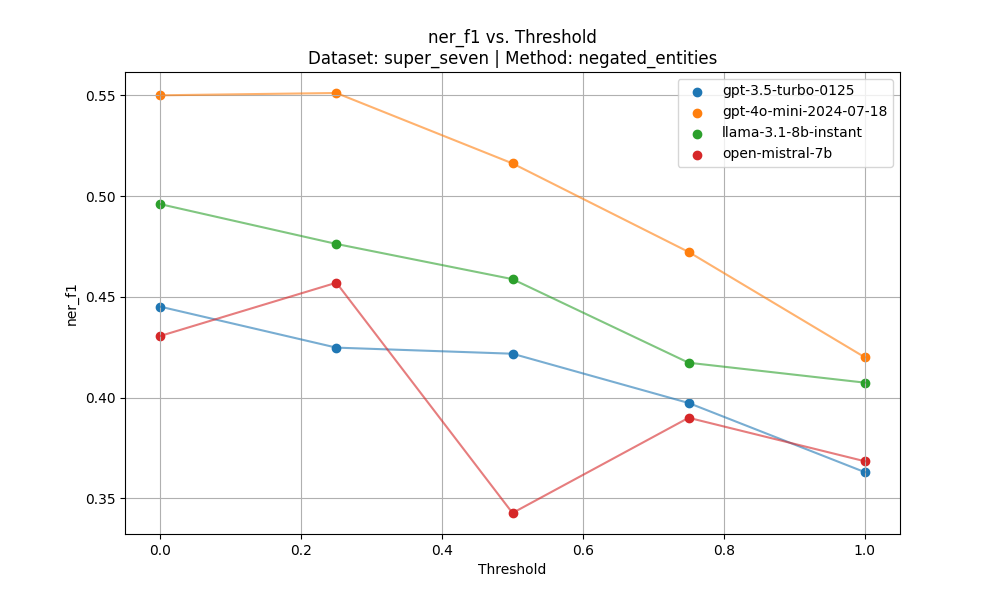}
        \caption{Negated Entities}
    \end{subfigure}
    \vskip\baselineskip
    \begin{subfigure}[b]{0.45\textwidth}
        \includegraphics[width=\linewidth]{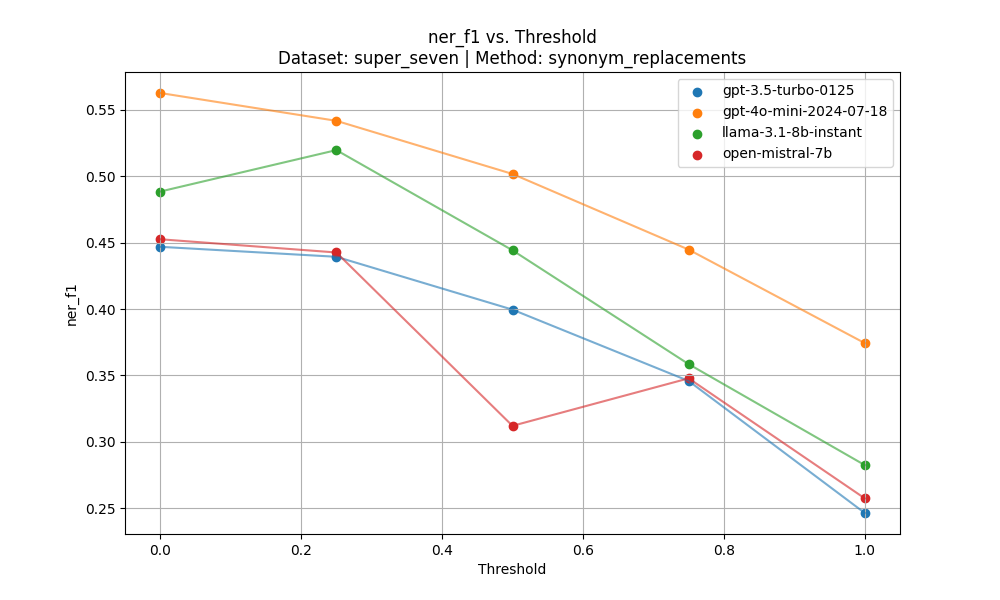}
        \caption{Synonym Replacements}
    \end{subfigure}
    \hfill
    \begin{subfigure}[b]{0.45\textwidth}
        \includegraphics[width=\linewidth]{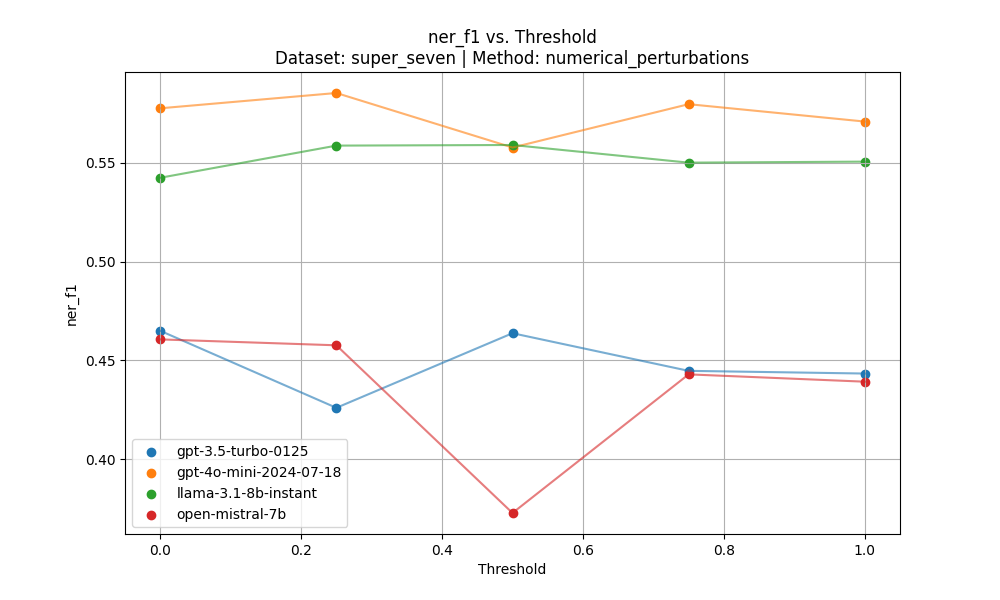}
        \caption{Numerical Perturbations}
    \end{subfigure}
    \Description{Line plots showing NER F1 degradation across perturbation thresholds for masked, negated, synonym, and numerical changes. Masking and synonym replacement cause sharp declines, while numerical perturbations have little effect.}
    \caption{Entity-level recognition performance degrades under perturbation, particularly with masking and synonym edits. This figure shows how NER F1 scores drop as perturbation intensity increases across four types. Masked and synonym-modified inputs reduce clinical entity recall most significantly.}
\end{figure*}

\subsubsection{ROUGE-L vs. Threshold}

\begin{figure*}[H]
    \centering
    \begin{subfigure}[b]{0.45\textwidth}
        \includegraphics[width=\linewidth]{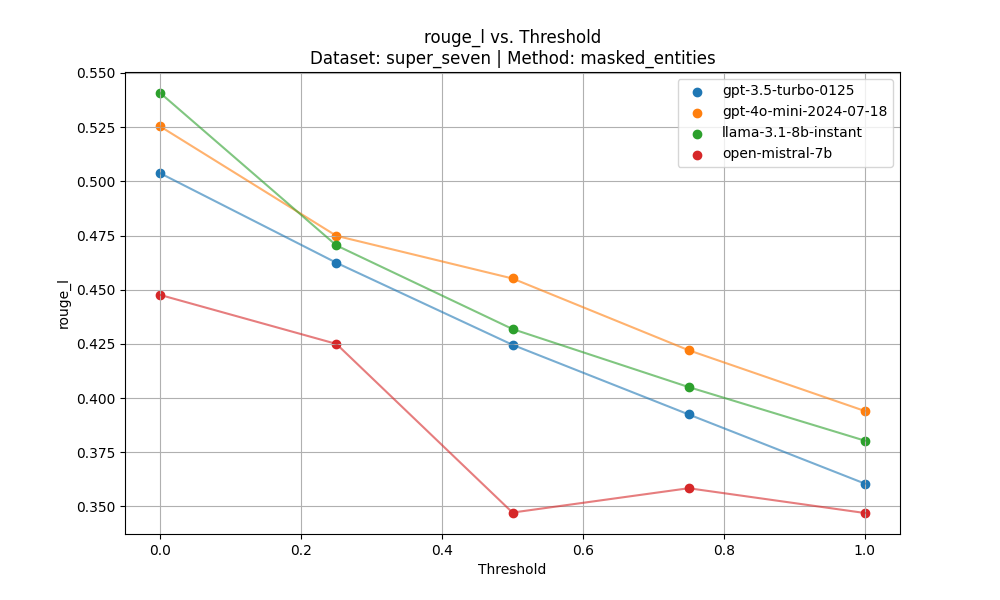}
        \caption{Masked Entities}
    \end{subfigure}
    \hfill
    \begin{subfigure}[b]{0.45\textwidth}
        \includegraphics[width=\linewidth]{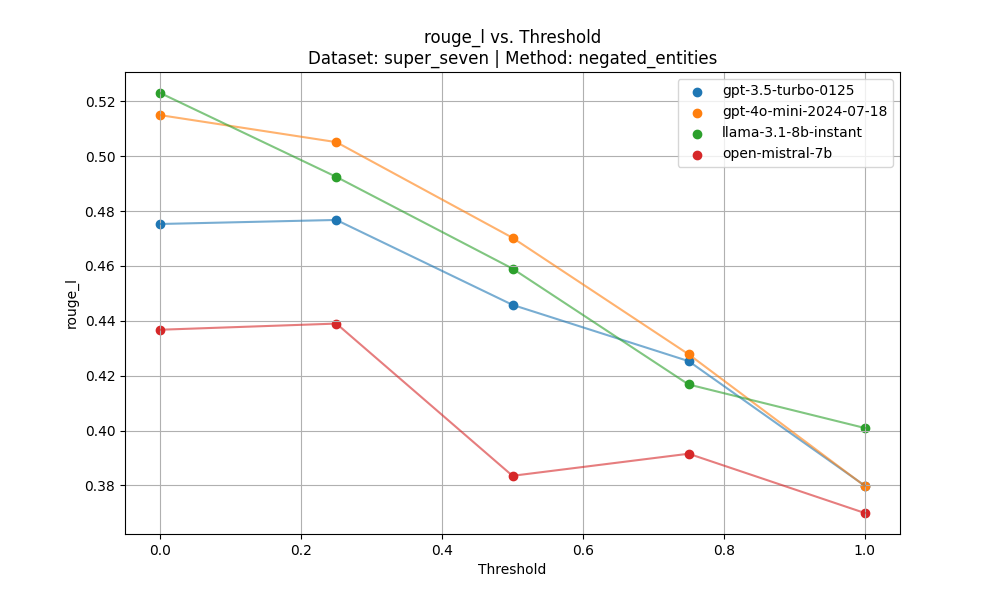}
        \caption{Negated Entities}
    \end{subfigure}
    \vskip\baselineskip
    \begin{subfigure}[b]{0.45\textwidth}
        \includegraphics[width=\linewidth]{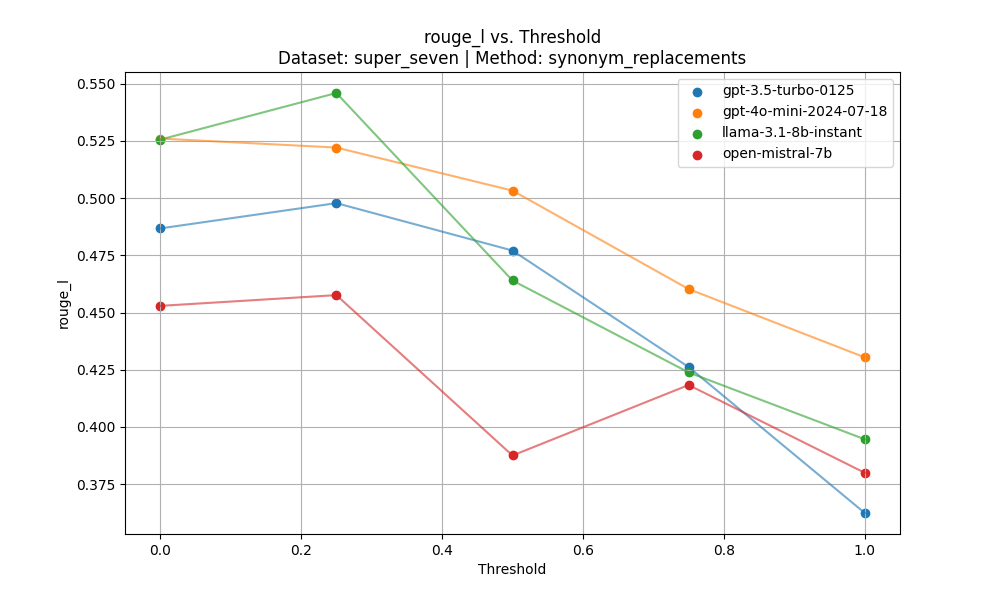}
        \caption{Synonym Replacements}
    \end{subfigure}
    \hfill
    \begin{subfigure}[b]{0.45\textwidth}
        \includegraphics[width=\linewidth]{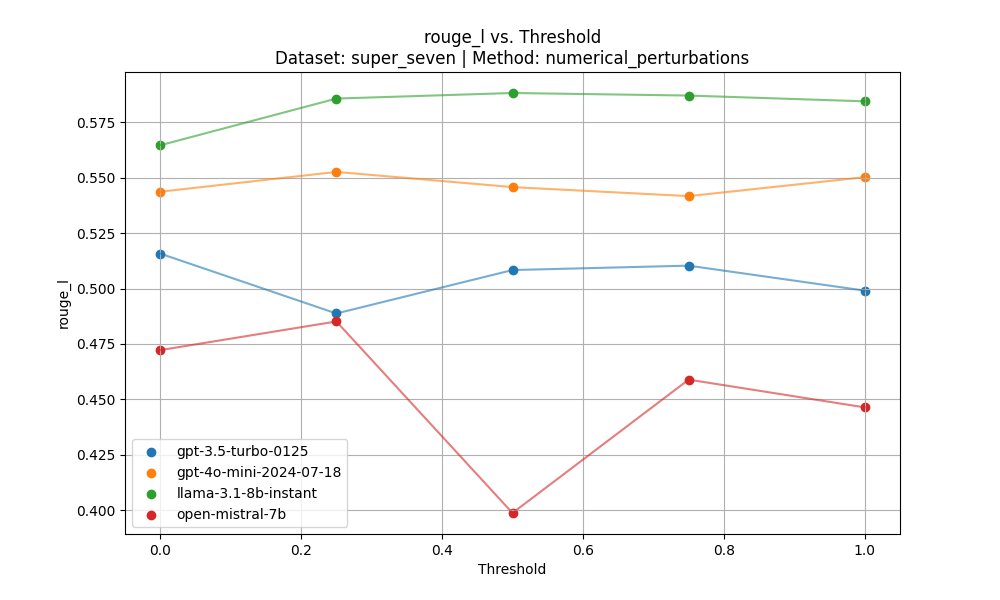}
        \caption{Numerical Perturbations}
    \end{subfigure}
    \Description{Four line plots showing ROUGE-L scores degrading across perturbation thresholds from 0 to 100 percent for masked, negated, synonym, and numerical variants. Masked and synonym perturbations show the steepest declines, especially in earlier thresholds.}
    \caption{ROUGE-L scores degrade with increasing perturbation intensity, reflecting reduced content-level overlap between original and perturbed notes. Masked entities and synonym replacements consistently produce the largest drops, highlighting their disruptive impact on lexical structure.}
\end{figure*}

\subsubsection{NER Jaccard vs. Threshold}

\begin{figure*}[H]
    \centering
    \begin{subfigure}[b]{0.45\textwidth}
        \includegraphics[width=\linewidth]{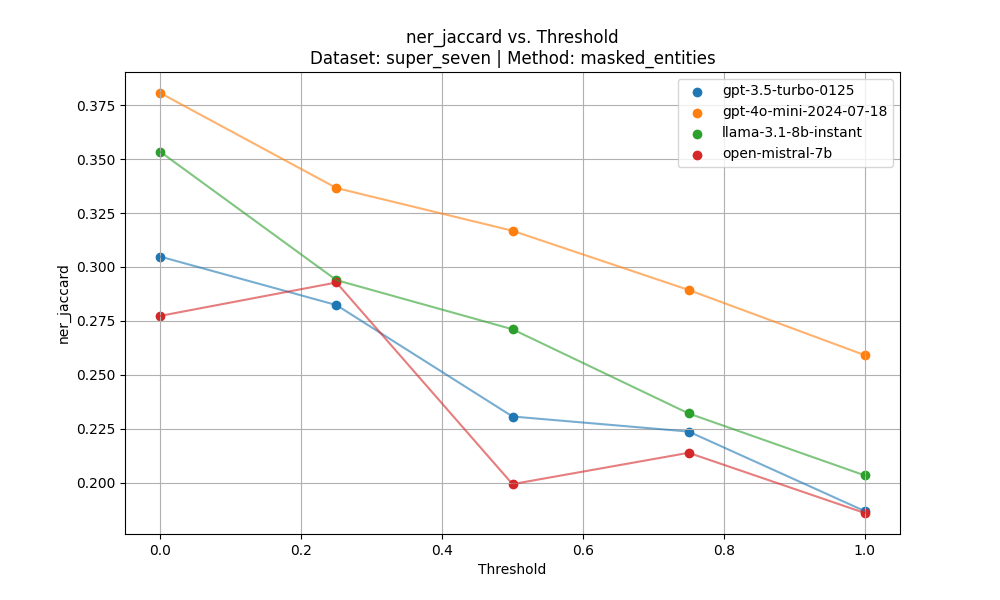}
        \caption{Masked Entities}
    \end{subfigure}
    \hfill
    \begin{subfigure}[b]{0.45\textwidth}
        \includegraphics[width=\linewidth]{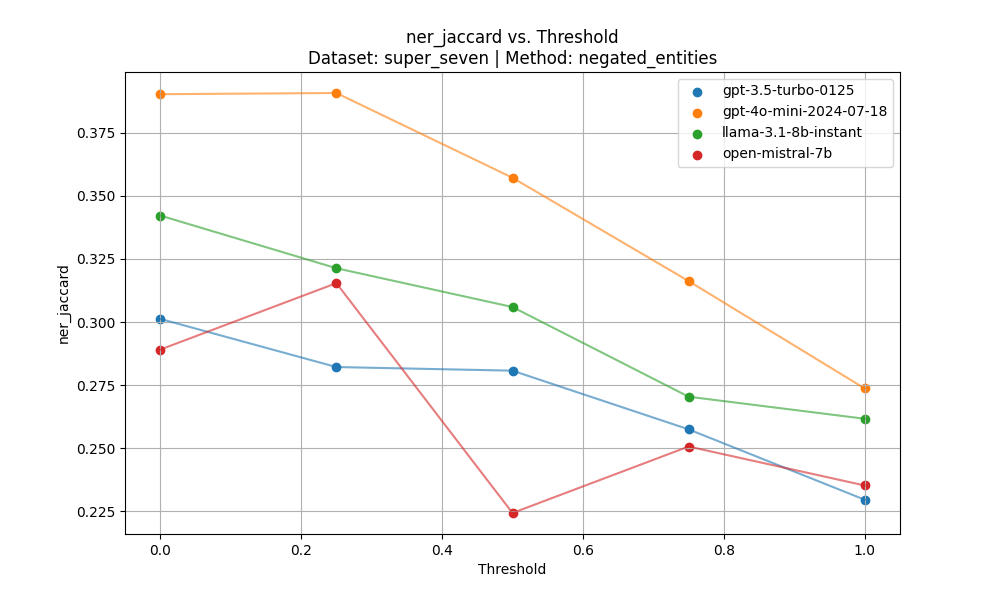}
        \caption{Negated Entities}
    \end{subfigure}
    \vskip\baselineskip
    \begin{subfigure}[b]{0.45\textwidth}
        \includegraphics[width=\linewidth]{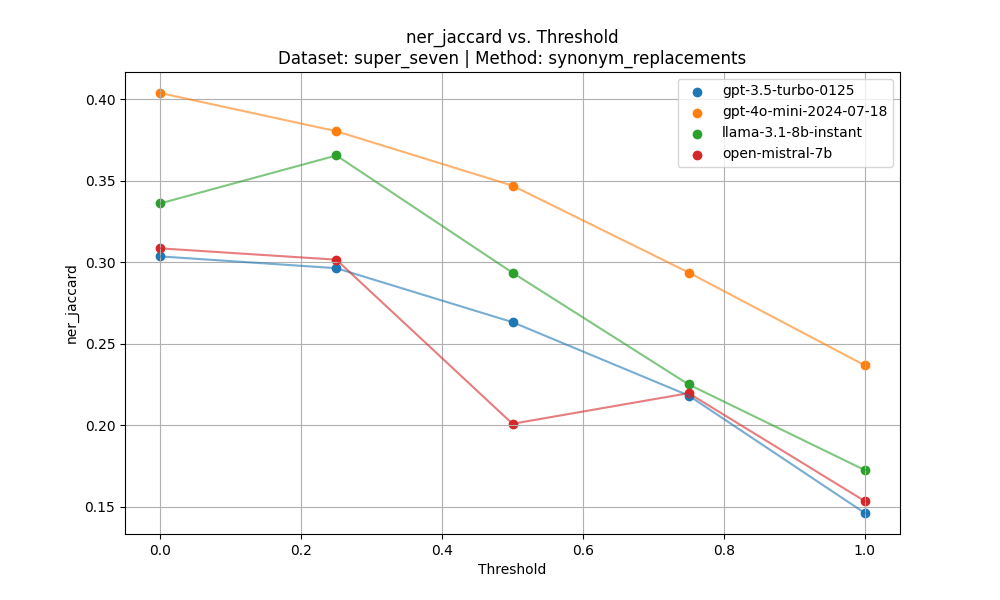}
        \caption{Synonym Replacements}
    \end{subfigure}
    \hfill
    \begin{subfigure}[b]{0.45\textwidth}
        \includegraphics[width=\linewidth]{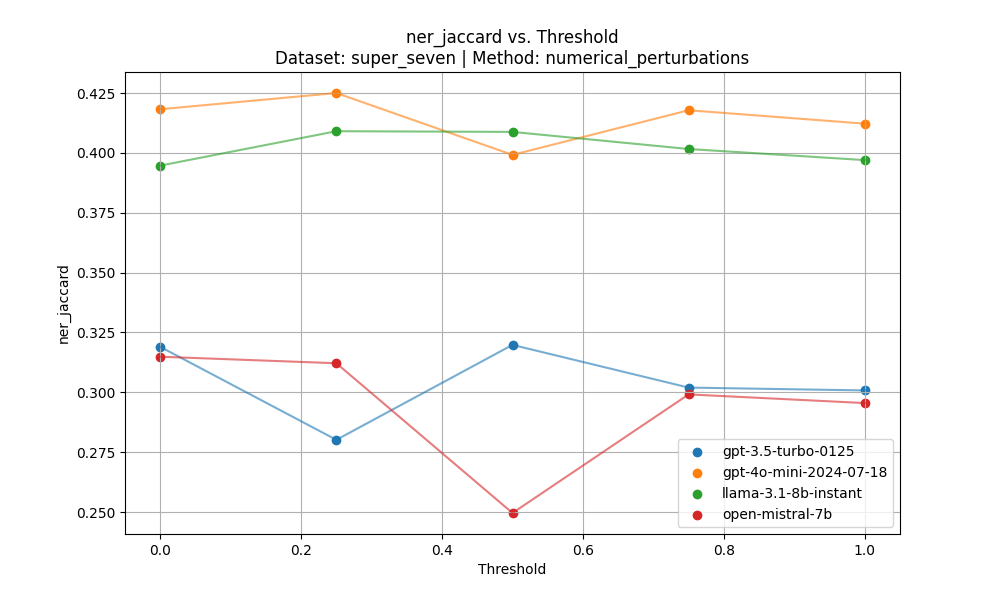}
        \caption{Numerical Perturbations}
    \end{subfigure}
    \Description{Four line plots showing how NER Jaccard scores degrade across perturbation thresholds for masked, negated, synonym, and numerical types. Masked and synonym edits cause steep drops, similar to NER F1 trends, indicating disruption in entity recognition consistency.}
    \caption{NER Jaccard degradation highlights entity drift under perturbation. These plots show how entity span overlap deteriorates as perturbation severity increases. Patterns closely mirror NER F1 degradation, reinforcing that semantic entity loss—especially under masking and synonym edits—is a key driver of robustness failure.}
\end{figure*}

\subsection{Hyperparameters for Geometric Analysis}
\label{appendix:hyperparams}
\subsubsection{Embedding Extraction}
\begin{itemize}
    \item \textbf{Embedding Model:} ClinicalBERT (pretrained, frozen)
    \item \textbf{Tokenization:} BertTokenizer (max length = 512 tokens)
    \item \textbf{Embedding Representation:} Mean pooling over last hidden layer
    \item \textbf{Embedding Dimension:} 768
\end{itemize}

\subsubsection{PCA Reduction}
\begin{itemize}
    \item \textbf{Input:} Embeddings of unperturbed clinical notes (N = 100)
    \item \textbf{Normalization:} Zero-centered with unit variance per dimension
    \item \textbf{Components Retained:} 90\% explained variance (typically ~30-40 components)
    \item \textbf{Library Used:} scikit-learn PCA (randomized SVD solver)
\end{itemize}

\subsubsection{Latent-Space Classifier}
\begin{itemize}
    \item \textbf{Classifier:} Logistic Regression (1-vs-rest)
    \item \textbf{Training Set:} PCA-reduced embeddings of original (unperturbed) notes
    \item \textbf{Labels:} Ground truth diagnoses (7-class classification)
    \item \textbf{Regularization:} L2, inverse regularization strength $C=10.0$
    \item \textbf{Solver:} lgfbs
    \item \textbf{Max Iteration}: 1000
    \item \textbf{Cross-validation:} 10-fold 
\end{itemize}

\subsection{Illustrated Example}

The following example illustrates the flow of clinical data for a case with GERD as diagnosis. We illustrate the case of Masked Perturbation at 100\%. The following entities will be masked during the perturbation.

We illustrate the Clinical Note with Fig~\ref{fig:gerd_note}, Forward Diagnosis for this case with Fig~\ref{fig:gerd_f_dia}, backward diagnosis with Fig~\ref{fig:gerd_b_dia} and final diagnostic reasoning with Fig~\ref{fig:gerd_a_dia}. 

The masked entities perturbation case is illustrated with Fig~\ref{fig:gerd_note_masked}, forward reasoning with Fig~\ref{fig:gerd_f_dia_masked}, backward reasoning with Fig~\ref{fig:gerd_b_dia_masked}, final reasoning with Fig~\ref{fig:gerd_a_dia_masked}. 

\textbf{chronic sinusitis, persistent cough, family history of asthma, 26-year-old male, hospitalizations, respiratory distress, asthma attacks, bronchodilator, wheezing, common allergies}

\begin{figure}[htbp]
\centering
\begin{minipage}{0.9\linewidth}
\begin{lstlisting}
"Patient Details:
Age: 57
Sex: Female

Chief Complaint:
The patient reports experiencing symptoms that worsen after eating.

History of Present Illness:
The patient describes her pain as haunting, sensitive, a knife stroke, sickening, and scary. The pain is located in the lower chest, upper chest, hypochondrium (left), and epigastric regions. The intensity of the pain is unknown, and it radiates to the upper chest. The pain's onset and duration are unknown, but it is alleviated while sitting up. The patient also reports a burning sensation that starts in the stomach and goes up into the throat, associated with a bitter taste in the mouth.

Past Medical History:
The patient has a hiatal hernia and asthma, for which she has used a bronchodilator in the past. She is currently pregnant.

Medications and Allergies:
No current medications were reported, but the patient has a history of using a bronchodilator for asthma. She has no known allergies.

Physical Examination:
Vitals: Not provided in the data
General Appearance: Not provided in the data
Pertinent Exam Findings: The patient reports pain in various chest regions, and a burning sensation with a bitter taste in the mouth. The pain is alleviated while sitting up. No further physical examination findings were provided in the data.

Additional Notes:
The patient also reports being significantly overweight compared to people of the same height, drinking alcohol excessively, and smoking cigarettes. She has not traveled out of the country in the last 4 weeks. These factors may be relevant to the patient's current symptoms and should be considered in the diagnostic and management process."
\end{lstlisting}
\end{minipage}
\caption{Clinical Note for diagnosis GERD}
\label{fig:gerd_note}
\end{figure}

\begin{figure}[htbp]
\centering
\begin{minipage}{0.9\linewidth}
\begin{lstlisting}
"1. The patient is a 57-year-old female.
2. She reports symptoms worsening after eating.
3. The pain is described as haunting, sensitive, a knife stroke, sickening, and scary.
4. The pain is located in the lower chest, upper chest, hypochondrium (left), and epigastric regions.
5. The pain radiates to the upper chest.
6. The intensity and duration of the pain are unknown.
7. The pain is alleviated while sitting up.
8. A burning sensation is reported, starting in the stomach and going up into the throat.
9. The patient has a bitter taste in the mouth.
10. She has a history of hiatal hernia and asthma.
11. She has used a bronchodilator for asthma in the past.
12. She is currently pregnant.
13. No current medications were reported.
14. She has no known allergies.
15. The patient reports being significantly overweight.
16. She drinks alcohol excessively.
17. She smokes cigarettes.
18. No further physical examination findings were provided.
19. The patient has not traveled out of the country in the last 4 weeks.
20. These factors (overweight, alcohol consumption, smoking) may be relevant to the patient's current symptoms."
\end{lstlisting}
\end{minipage}
\caption{Forward reasoning for GERD}
\label{fig:gerd_f_dia}
\end{figure}

\begin{figure}[htbp]
\centering
\begin{minipage}{0.9\linewidth}
\begin{lstlisting}
"1. Patient reports symptoms worsening after eating. (Observation)
2. Patient describes pain as haunting, sensitive, and radiating to the upper chest. (Observation)
3. Patient reports a burning sensation from stomach to throat with a bitter taste. (Observation)
4. Patient has a hiatal hernia. (Past Medical History)
5. Symptoms of GERD (gastroesophageal reflux disease) include heartburn, regurgitation, and pain that worsens after eating. (Knowledge of GERD)
6. GERD symptoms can be exacerbated by hiatal hernia. (Knowledge of GERD and hiatal hernia)
7. The patient's symptoms, including pain, burning sensation, and bitter taste, are consistent with GERD. (Logical deduction)
8. The patient's history of hiatal hernia, along with her symptoms, increases the likelihood of GERD. (Logical deduction)
9. The patient's symptoms are not typical of other conditions, such as heart disease or gallbladder disease, given her age and lack of other risk factors. (Logical deduction)
10. Therefore, the most likely diagnosis is GERD. (Final deduction)"
\end{lstlisting}
\end{minipage}
\caption{Backward reasoning for GERD}
\label{fig:gerd_b_dia}
\end{figure}

\begin{figure}[htbp]
\centering
\begin{minipage}{0.9\linewidth}
\begin{lstlisting}
"The forward and backward reasoning chains are consistent in supporting the diagnosis of GERD (gastroesophageal reflux disease).

Observations:
1. Symptoms worsen after eating.
2. Pain described as haunting, sensitive, and radiating to the upper chest.
3. Burning sensation from stomach to throat with a bitter taste.

Logical Deductions:
1. Patient's symptoms are consistent with GERD.
2. Patient's history of hiatal hernia increases the likelihood of GERD.
3. Patient's symptoms are not typical of other conditions, such as heart disease or gallbladder disease.
4. The patient's lifestyle factors (overweight, alcohol consumption, smoking) may be relevant to her symptoms.
5. The patient's symptoms are alleviated while sitting up, which is a common symptom of GERD.
6. The patient's use of a bronchodilator for asthma does not directly contribute to the diagnosis of GERD, but it does not rule it out either.
7. The patient's current pregnancy may not directly affect the diagnosis of"
\end{lstlisting}
\end{minipage}
\caption{Aggregated reasoning for GERD}
\label{fig:gerd_a_dia}
\end{figure}

\begin{figure}[htbp]
\centering
\begin{minipage}{0.9\linewidth}
\begin{lstlisting}
"Patient Details:
Age: 57
Sex: [MASK]

Chief Complaint:
The patient reports experiencing [MASK] that worsen after eating.

History of Present Illness:
The patient describes her [MASK] as haunting, sensitive, a knife stroke, sickening, and scary. The [MASK] is located in the [MASK], [MASK], hypochondrium (left), and [MASK]. The intensity of the [MASK] is unknown, and it radiates to the [MASK]. The [MASK]'s onset and duration are unknown, but it is alleviated while sitting up. The patient also reports a [MASK] that starts in the [MASK] and goes up into the [MASK], associated with a [MASK] in the mouth.

Past Medical History:
The patient has a [MASK] and [MASK], for which she has used a [MASK] in the past. She is currently [MASK].

Medications and Allergies:
No current medications were reported, but the patient has a history of using a [MASK] for [MASK]. She has no known allergies.

Physical Examination:
Vitals: Not provided in the data
General Appearance: Not provided in the data
Pertinent Exam Findings: The patient reports [MASK] in various chest regions, and a [MASK] with a [MASK] in the mouth. The [MASK] is alleviated while sitting up. No further physical examination findings were provided in the data.

Additional Notes:
The patient also reports being significantly [MASK] compared to people of the same height, drinking [MASK] excessively, and [MASK]. She has not traveled out of the country in the last 4 weeks. These factors may be relevant to the patient's current [MASK] and should be considered in the diagnostic and management process."
\end{lstlisting}
\end{minipage}
\caption{Masked Clinical Note for diagnosis GERD}
\label{fig:gerd_note_masked}
\end{figure}

\begin{figure}[htbp]
\centering
\begin{minipage}{0.9\linewidth}
\begin{lstlisting}
"1. Patient is a 57-year-old [MASK] with a chief complaint of unexplained discomfort worsening after eating.
2. The discomfort is described as haunting, sensitive, a knife stroke, sickening, and scary.
3. The discomfort is located in the epigastric region, left hypochondrium, and umbilical region.
4. The intensity and duration of the discomfort are unknown, but it radiates to the back.
5. The discomfort is alleviated while sitting up.
6. The patient also reports a burning sensation starting in the throat and going up into the mouth, associated with a sensation in the mouth.
7. The patient has a history of hypertension and diabetes, and has used a diuretic in the past.
8. No current medications were reported, but the patient has a history of using a diuretic for hypertension.
9. The patient has no known allergies.
10. The patient reports feeling significantly underweight, drinking alcohol excessively, and having a poor appetite.
11. No further physical examination findings were provided.
12. The patient has not traveled out of the country in the last 4 weeks.
13. These factors may be relevant to the patient's current symptoms and should be considered in the diagnostic and management process."
\end{lstlisting}
\end{minipage}
\caption{Forward reasoning for Masked Perturbation of 100\% GERD}
\label{fig:gerd_f_dia_masked}
\end{figure}

\begin{figure}[htbp]
\centering
\begin{minipage}{0.9\linewidth}
\begin{lstlisting}
"1. The patient reports symptoms of heartburn that worsen after eating.
2. The heartburn is described as haunting, sensitive, and radiating to the left hypochondrium and back.
3. The patient has a history of GERD and has used a proton pump inhibitor (PPI) in the past.
4. The patient's symptoms are alleviated while sitting up, which is a common symptom of GERD.
5. The patient's excessive alcohol consumption and obesity may contribute to the worsening of GERD symptoms.
6. No physical examination findings suggest an alternative diagnosis, such as a cardiac issue.

Logical deductions:
1. The patient's symptoms of heartburn, worsened by eating and alleviated by sitting up, are consistent with GERD.
2. The patient's history of GERD and use of a PPI support the diagnosis of GERD.
3. The patient's obesity and excessive alcohol consumption may exacerbate GERD symptoms.
4. The absence of physical examination findings suggesting an alternative diagnosis supports the diagnosis of GERD."
\end{lstlisting}
\end{minipage}
\caption{Backward reasoning for Masked Perturbation of 100\% GERD}
\label{fig:gerd_b_dia_masked}
\end{figure}

\begin{figure}[htbp]
\centering
\begin{minipage}{0.9\linewidth}
\begin{lstlisting}
"The forward and backward reasoning chains are consistent in supporting the diagnosis of GERD.

Observations:
1. Patient's discomfort worsens after eating and is alleviated while sitting up.
2. Discomfort is described as haunting, sensitive, and radiates to the left hypochondrium and back.
3. Patient has a history of hypertension, diabetes, and past use of a diuretic.
4. Patient reports excessive alcohol consumption, poor appetite, and feeling significantly underweight.

Logical deductions:
1. Symptoms are consistent with GERD.
2. Patient's history of GERD and use of a PPI support the diagnosis.
3. Patient's lifestyle factors may exacerbate GERD symptoms.
4. Absence of physical examination findings suggesting an alternative diagnosis supports the diagnosis of GERD."
\end{lstlisting}
\end{minipage}
\caption{Final diagnosistic reasoning for Masked Perturbation of 100\% GERD}
\label{fig:gerd_a_dia_masked}
\end{figure}

\subsection {Per-Dimension Variance Plots and Interpretation Synthetic Notes}
\label{appendix:dim_variance}
\begin{figure*}[htbp]
    \centering
    \begin{subfigure}[b]{0.5\linewidth}
        \includegraphics[width=\linewidth]{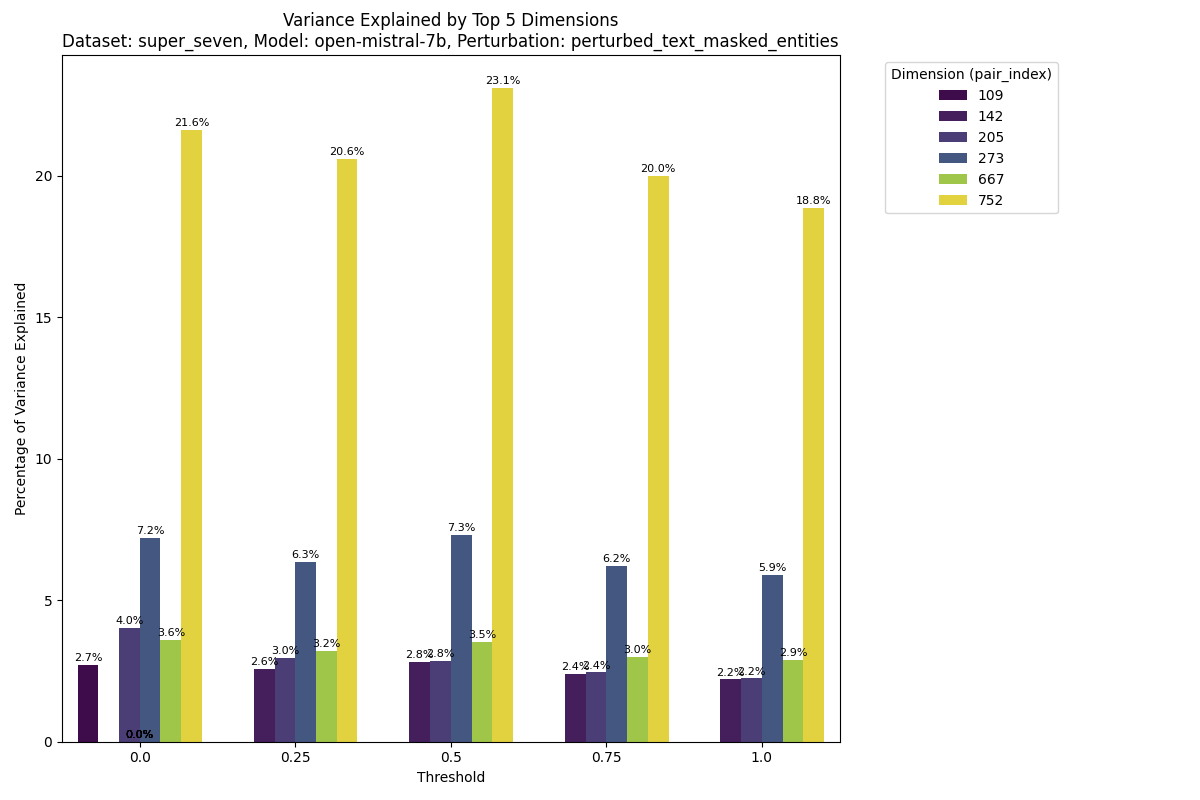}
        \caption{Masked Entities}
    \end{subfigure}%
    \begin{subfigure}[b]{0.5\linewidth}
        \includegraphics[width=\linewidth]{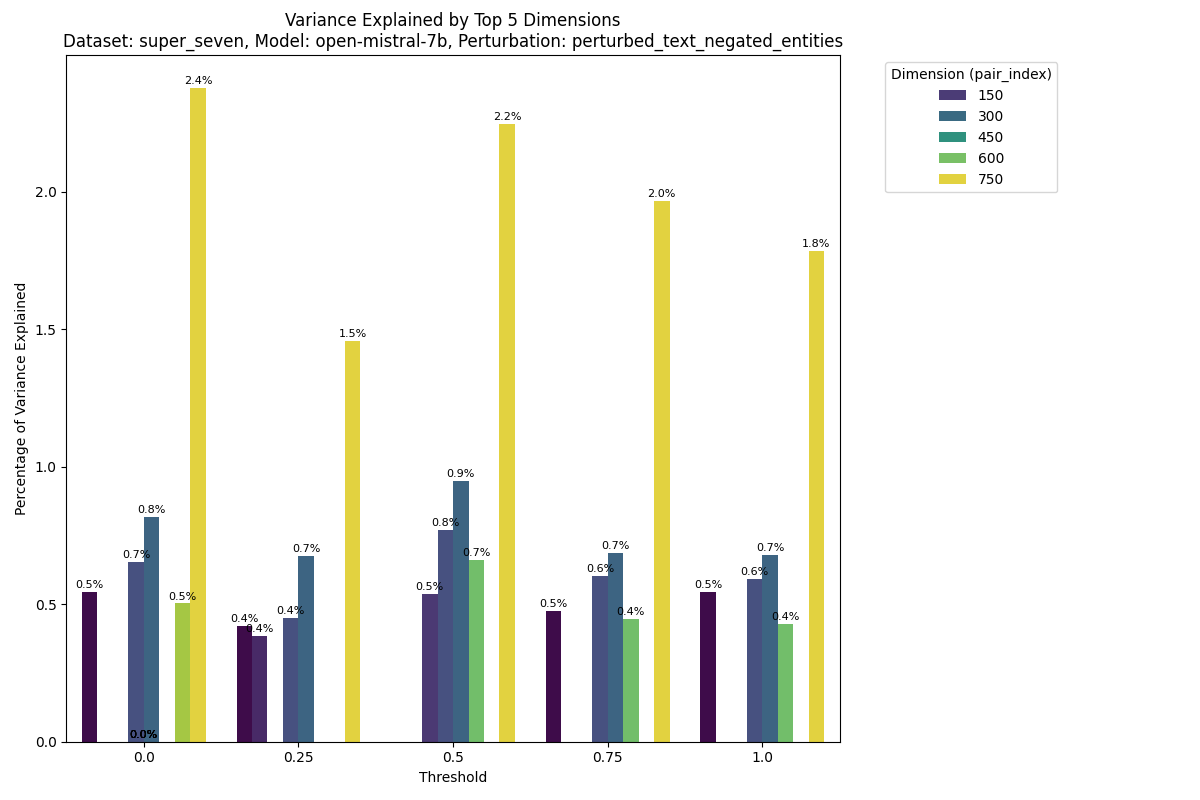}
        \caption{Negated Entities}
    \end{subfigure}%

    \begin{subfigure}[b]{0.5\linewidth}
        \includegraphics[width=\linewidth]{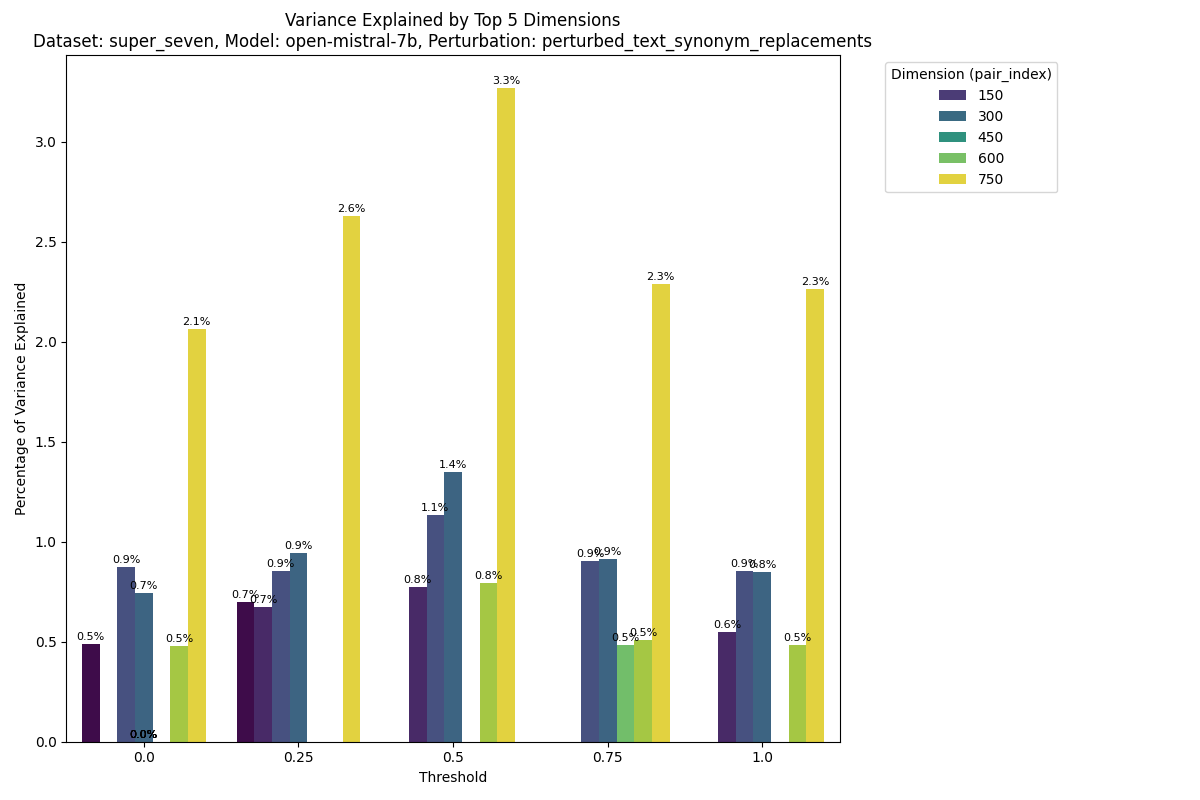}
        \caption{Synonym Replacements}
    \end{subfigure}%
    \begin{subfigure}[b]{0.5\linewidth}
        \includegraphics[width=\linewidth]{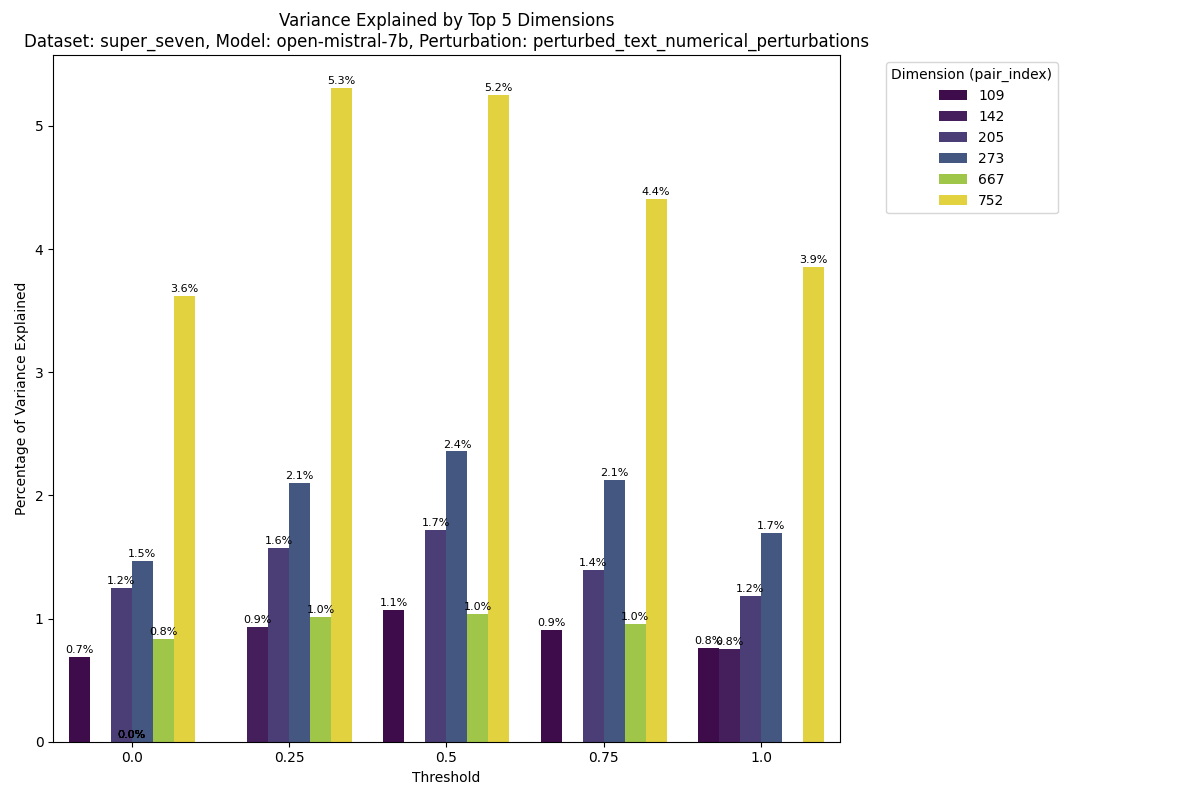}
        \caption{Numerical Perturbations}
    \end{subfigure}%
    \caption{Variance explained by top 5 latent dimensions for \texttt{open-mistral-7b} across perturbation types.}
    \label{fig:dim_variance_openmistral}
\end{figure*}

\begin{figure*}[htbp]
    \centering
    \begin{subfigure}[b]{0.5\linewidth}
        \includegraphics[width=\linewidth]{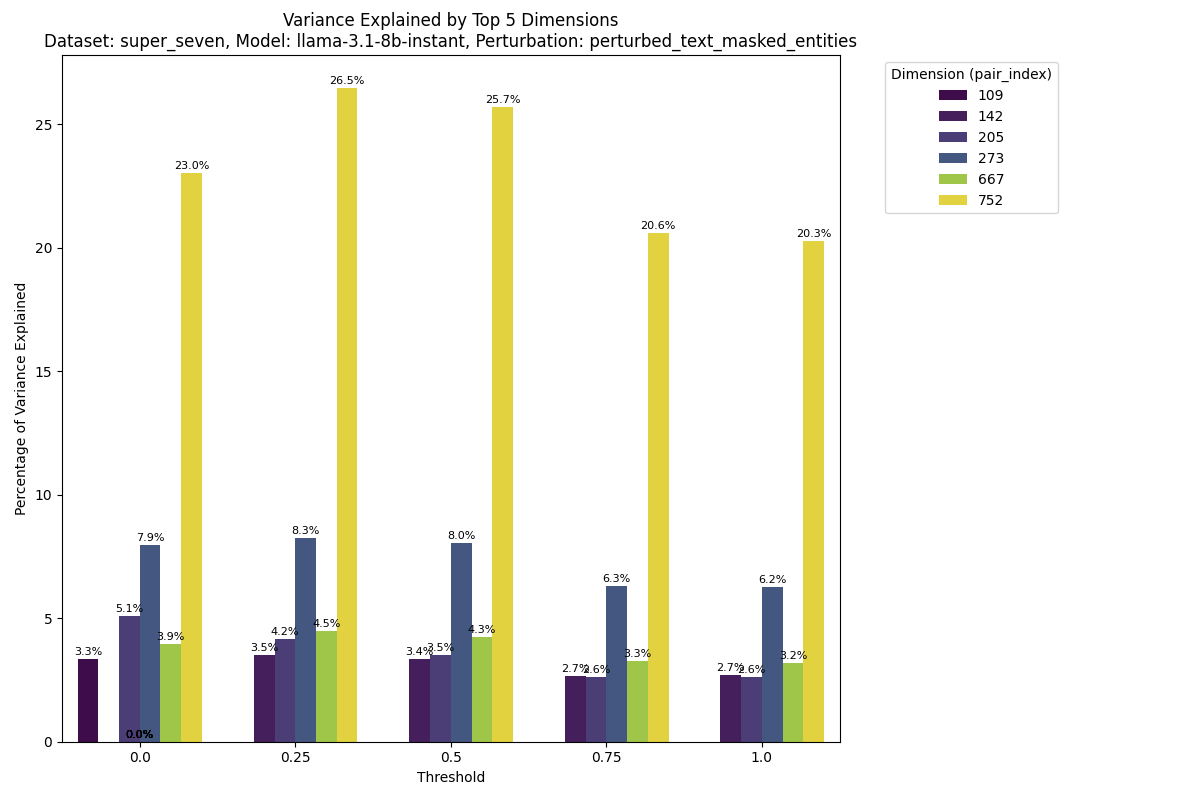}
        \caption{Masked Entities}
    \end{subfigure}%
    \begin{subfigure}[b]{0.45\linewidth}
        \includegraphics[width=\linewidth]{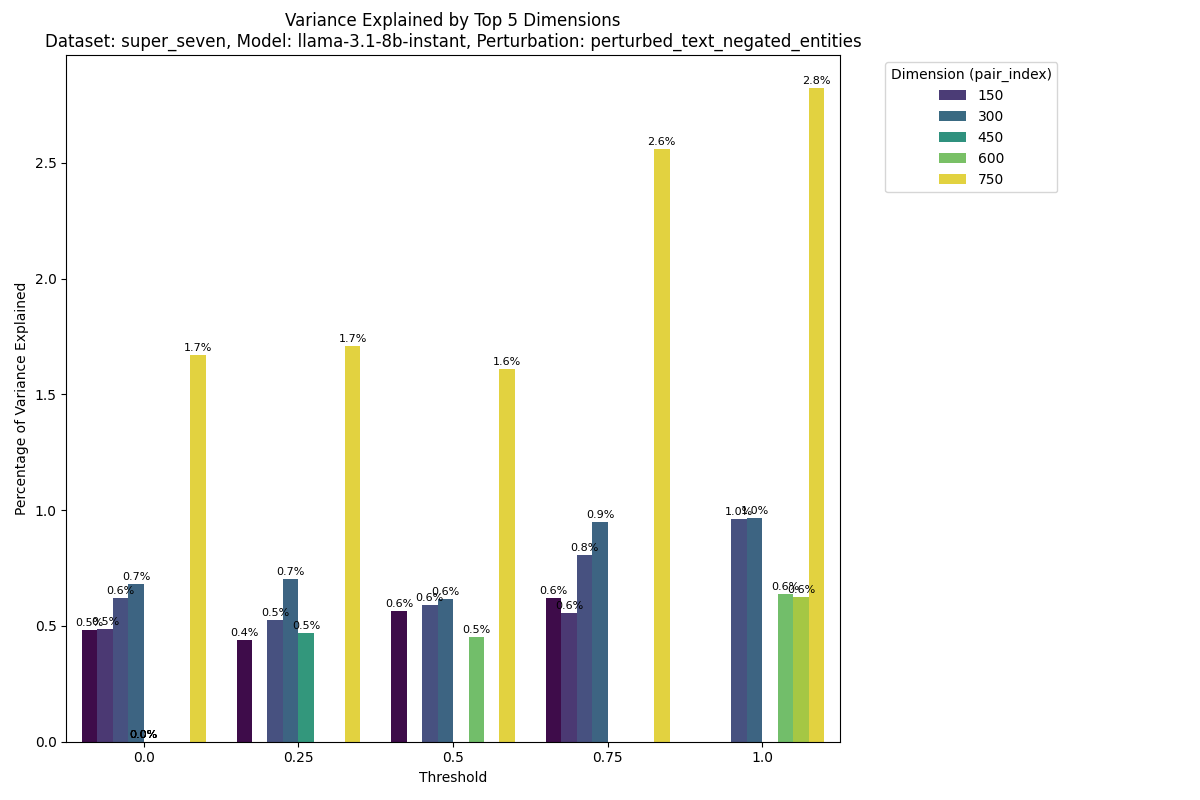}
        \caption{Negated Entities}
    \end{subfigure}%
    
    \begin{subfigure}[b]{0.5\linewidth}
        \includegraphics[width=\linewidth]{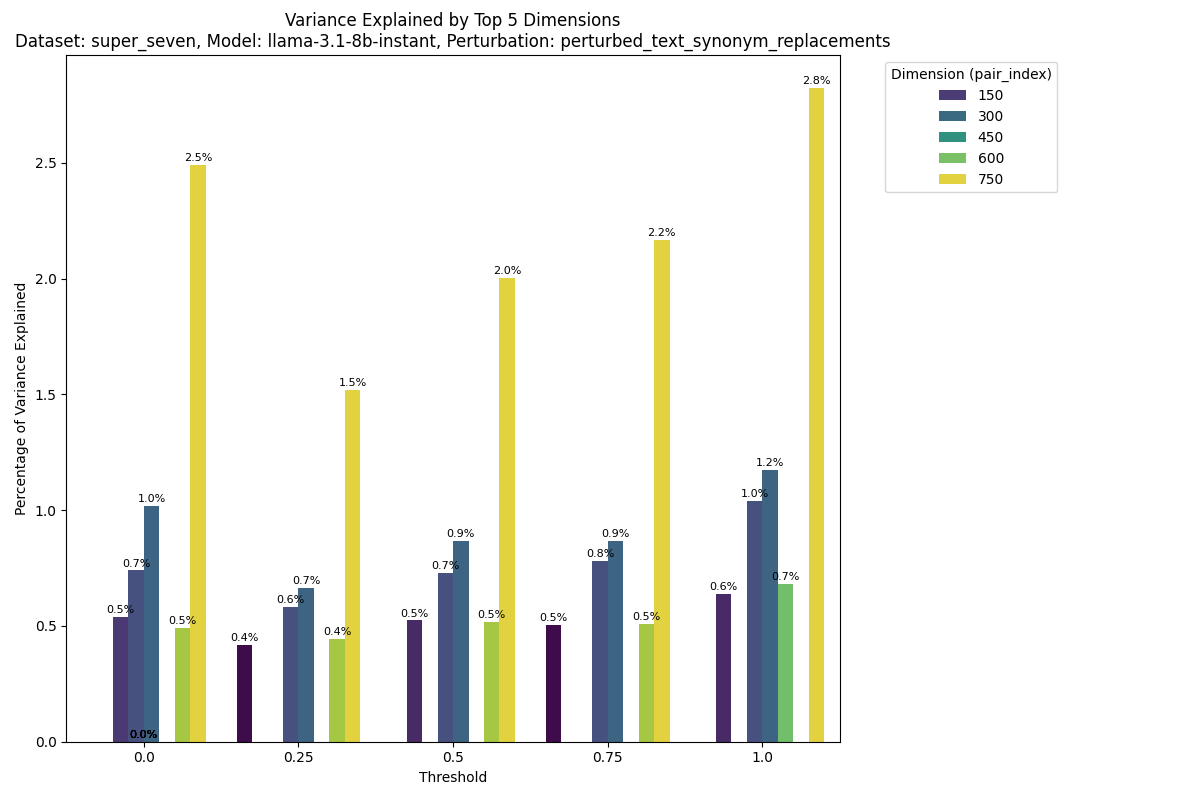}
        \caption{Synonym Replacements}
    \end{subfigure}%
    \begin{subfigure}[b]{0.5\linewidth}
        \includegraphics[width=\linewidth]{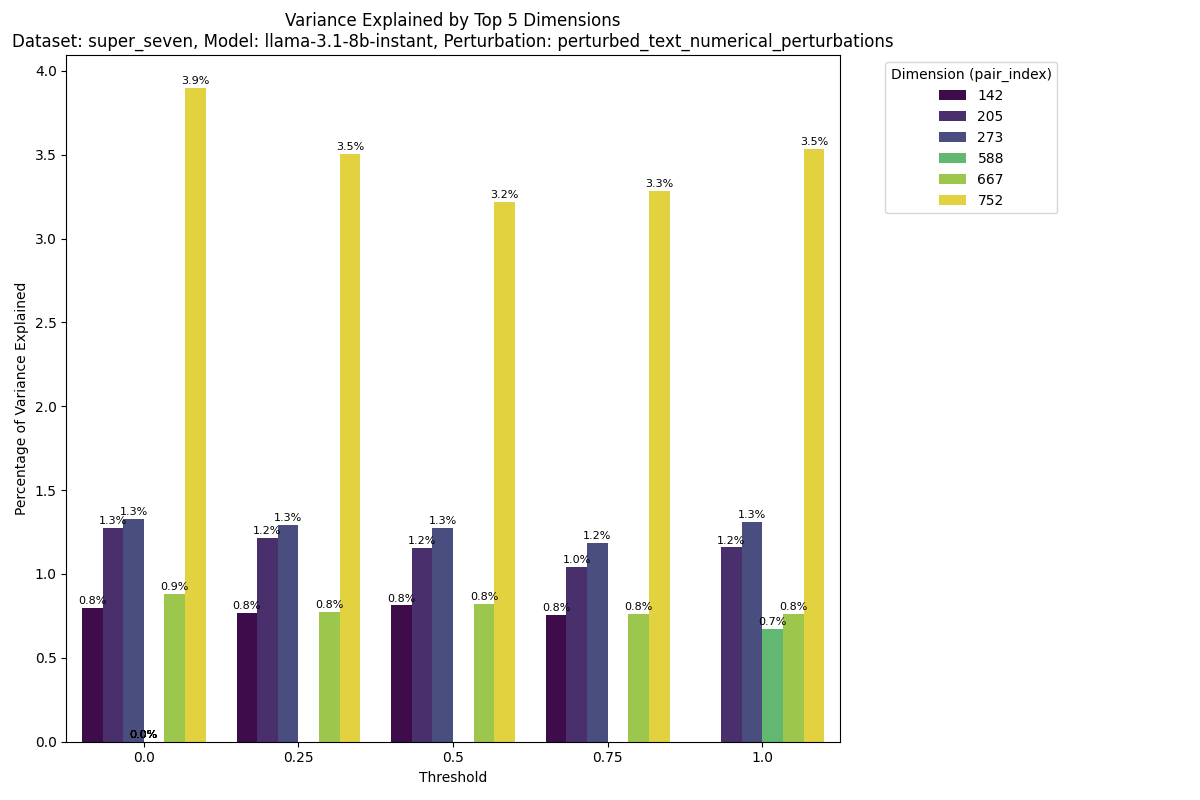}
        \caption{Numerical Perturbations}
    \end{subfigure}%
    \caption{Variance explained by top 5 latent dimensions for \texttt{lama\-3\.1\-8b\-instant} across perturbation types.}
    \label{fig:dim_variance_llama}
\end{figure*}

\begin{figure*}[htbp]
    \centering
    \begin{subfigure}[b]{0.5\linewidth}
        \includegraphics[width=\linewidth]{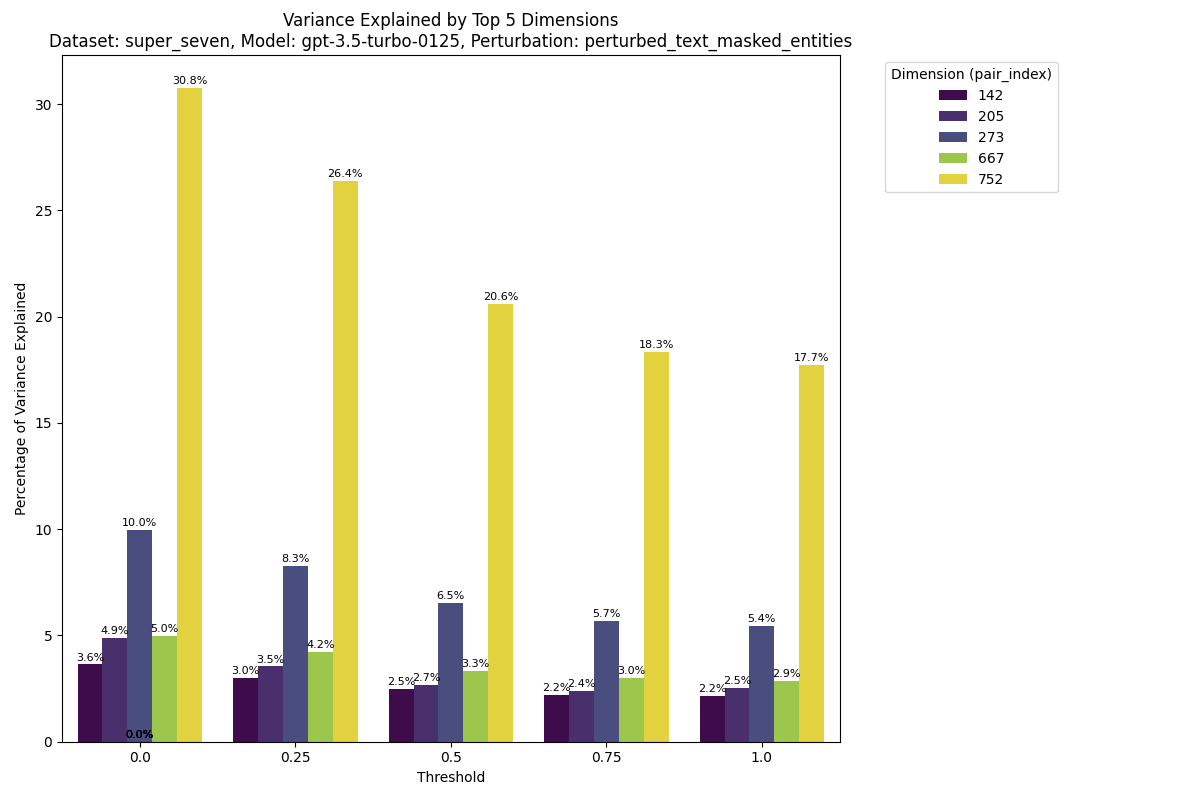}
        \caption{Masked Entities}
    \end{subfigure}%
    \begin{subfigure}[b]{0.5\linewidth}
        \includegraphics[width=\linewidth]{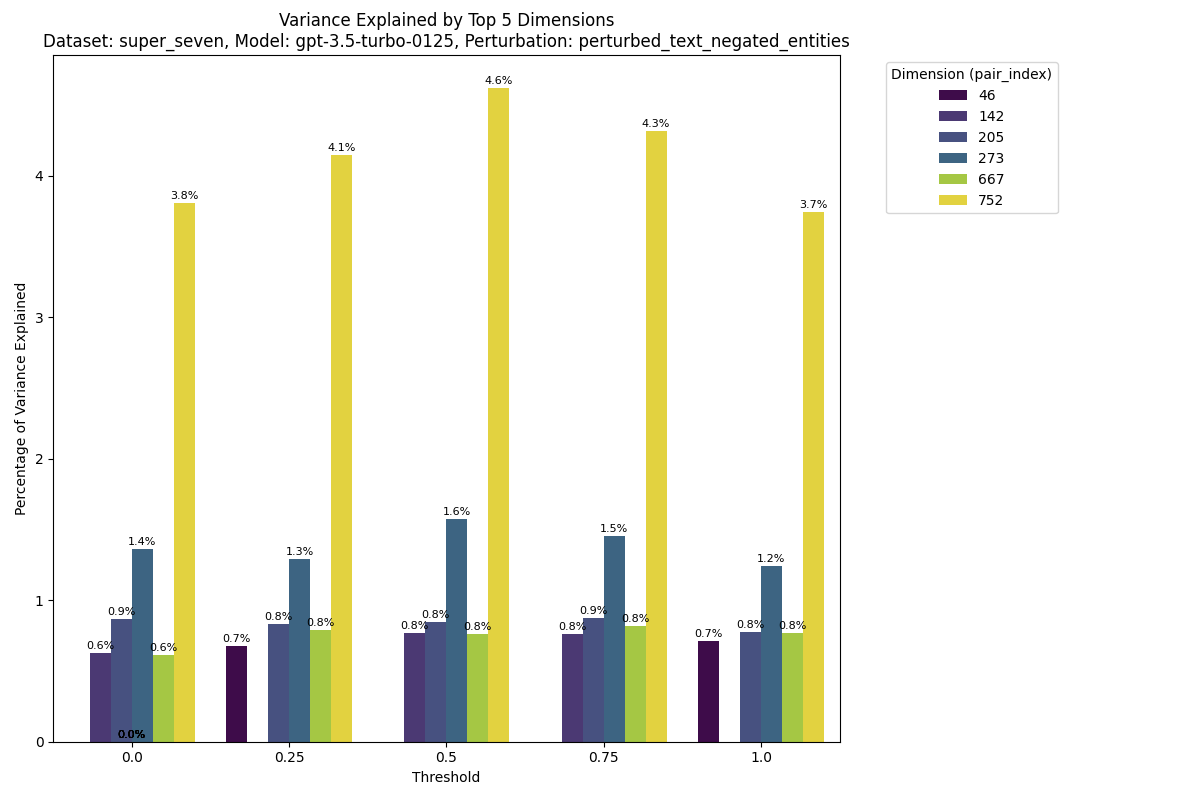}
        \caption{Negated Entities}
    \end{subfigure}%
    
    \begin{subfigure}[b]{0.5\linewidth}
        \includegraphics[width=\linewidth]{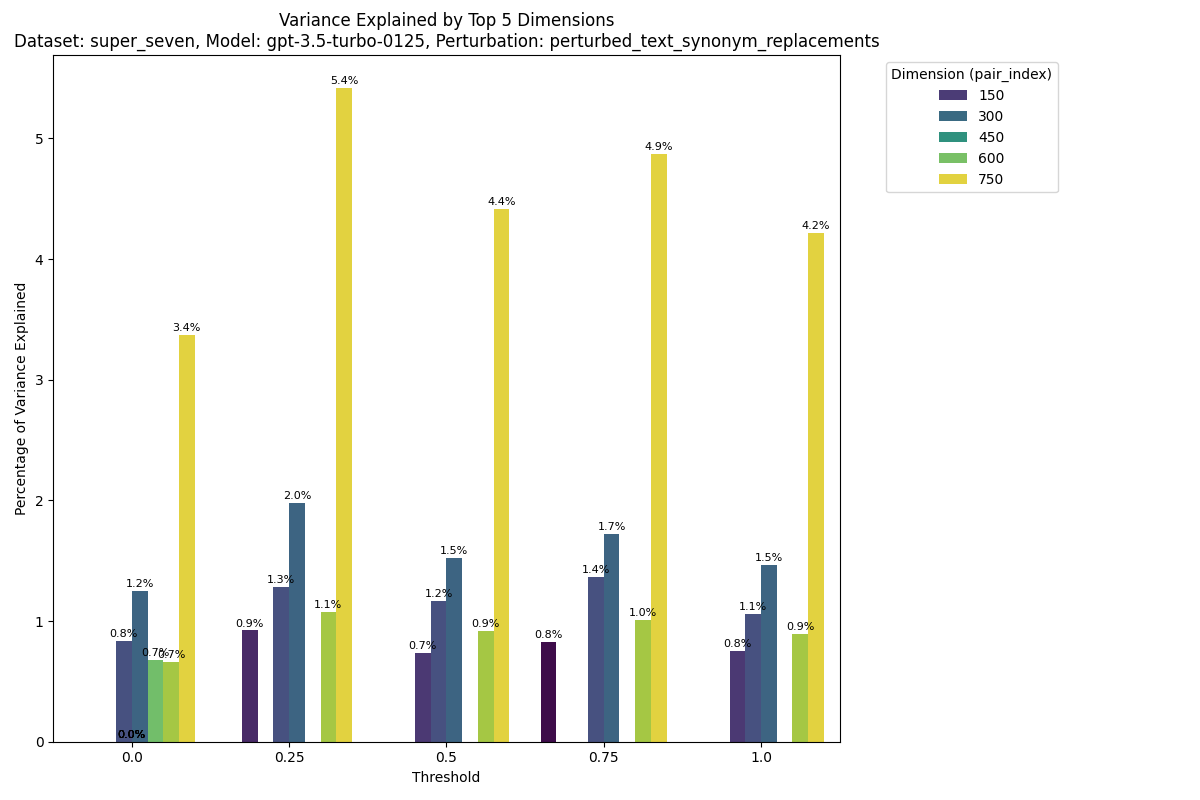}
        \caption{Synonym Replacements}
    \end{subfigure}%
    \begin{subfigure}[b]{0.5\linewidth}
        \includegraphics[width=\linewidth]{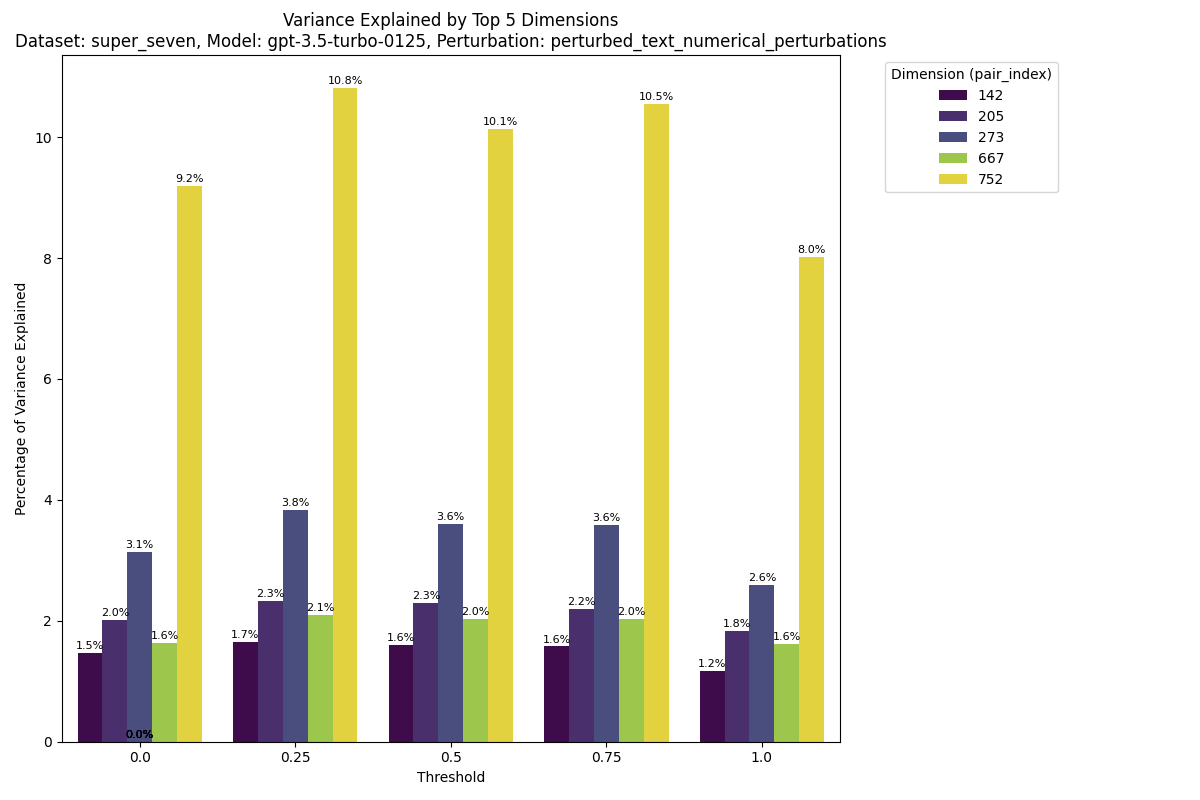}
        \caption{Numerical Perturbations}
    \end{subfigure}%
    \caption{Variance explained by top 5 latent dimensions for \texttt{gpt\-3\_5\-turbo\-0125} across perturbation types.}
    \label{fig:dim_variance_gpt-3-5}
\end{figure*}
To support our analysis in Section~\ref{sec:latent-collapse}, we present detailed variance decomposition plots for each model under all perturbation types. Each figure shows the top 5 latent dimensions with the highest explained variance across perturbation thresholds (0\%, 25\%, 50\%, 75\%, 100\%). Illustrated through Fig \ref{fig:dim_variance_llama}, Fig \ref{fig:dim_variance_openmistral} and Fig \ref{fig:dim_variance_gpt-3-5}.

\textbf{Key Insights}
\begin{itemize}
    \item Dimensions such as 273 and 752 consistently exhibit high sensitivity to masked and negated entity perturbations.
    \item GPT-3.5 and LLaMA display stronger activation along single dimensions, while GPT-4o exhibits more distributed variance across axes.
    \item Numerical perturbations exhibit the flattest distribution, indicating resilience in quantitative embeddings.
\end{itemize}

\subsubsection{Composite Visualization}

To reduce visual clutter, we aggregate per-model variance breakdowns into grouped figures by perturbation type. Full-resolution figures per model are available upon request.

\subsection{Elbow Plot of Latent Embedding Dimensionality}
\label{appendix:pca_elbow}
To select an appropriate dimensionality for LDFR evaluation, we performed Principal Component Analysis (PCA) on note embeddings from both synthetic and real clinical notes. The goal was to retain sufficient variance for meaningful diagnostic structure while avoiding overfitting or noise amplification in the latent classifier.

\begin{figure*}[htbp]
    \centering
    \includegraphics[width=0.95\linewidth]{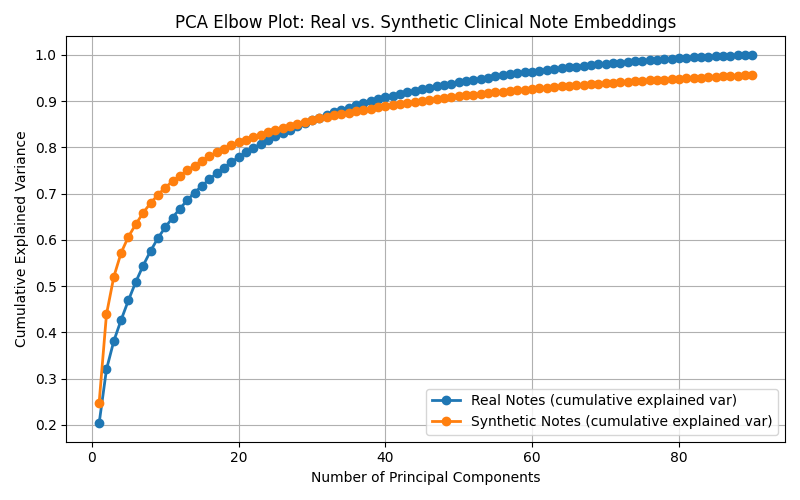}
    \caption{PCA Elbow Plot comparing cumulative explained variance for real (DiReCT) and synthetic clinical note embeddings. Real notes require more dimensions to capture 90\% of variance, reflecting their greater complexity and redundancy.}
    \label{fig:pca_elbow}
\end{figure*}

Figure~\ref{fig:pca_elbow} shows the cumulative explained variance as a function of the number of principal components.

For synthetic notes, 90\% of variance is typically captured within 30–35 components. In contrast, real clinical notes require approximately 45 components to reach the same threshold, consistent with their longer length and richer entity density.

We therefore fixed the PCA threshold at 90\% explained variance throughout our experiments. This ensures that the latent classifier operates in a stable, low-dimensional subspace that is both interpretable and aligned with clinical signal, as supported by the elbow point in both cases.


\appendix

\end{document}